\documentclass[10pt,journal,compsoc]{IEEEtran}
\usepackage[T1]{fontenc}
\usepackage[utf8]{inputenc}
\usepackage{babel}

\usepackage[pdftex]{graphicx}\usepackage{tikz}
\usepackage{comment}
\usepackage{amsmath,amssymb} 
\usepackage{color}
\usepackage{adjustbox}
\usepackage{algorithm}
\usepackage{algorithmic}
\ifCLASSOPTIONcompsoc
  \usepackage[nocompress]{cite}
\else
  \usepackage{cite}
\fi
\ifCLASSINFOpdf
\else
\fi
\begin{document}
%
\title{Object Tracking Using Spatio-Temporal Future Prediction}
%
%
%
%

\author{Yuan~Liu,
        Ruoteng~Li,
        Robby~T.~Tan,~\IEEEmembership{Member,~IEEE,}
        Yu~Cheng,
        Xiubao~Sui,~\IEEEmembership{Member,~IEEE} 
\IEEEcompsocitemizethanks{\IEEEcompsocthanksitem Y. Liu and X. Sui are
with School of Electronic and Optical Engineering, Nanjing University of Science
and Technology, Nanjing 210094, China. \protect\\
E-mail: walkyuan90@gmail.com, sxbhandsome@njust.edu.cn

\IEEEcompsocthanksitem R. T. Tan is with the Yale-NUS College and the Department of Electrical
and Computer Engineering, National University of Singapore, Singapore
119077. \protect\\
E-mail: robby.tan@nus.edu.sg.

\IEEEcompsocthanksitem R. Li and Y. Cheng are with the Department of Electrical and Computer Engineering,
National University of Singapore, Singapore 119077. \protect\\
E-mail: \{liruoteng, e0321276\}@u.nus.edu}}
\IEEEtitleabstractindextext{%
\begin{abstract}
Occlusion is a long-standing problem that causes many modern tracking methods to be erroneous. In this paper, we address the occlusion problem by exploiting the current and  future possible locations of the target object from its past trajectory.
To achieve this, we introduce a learning-based tracking method that takes into account background motion modeling and trajectory prediction. Our trajectory prediction module predicts the target object’s locations in the current and future frames based on the object's past trajectory.
Since, in the input video, the target object's trajectory is not only affected by the object  motion but also the camera motion, our background motion module estimates the camera motion, so that the object's trajectory can be made independent from it.
To dynamically switch between the appearance-based tracker and the trajectory prediction, we employ a network that can assess how good a tracking prediction is, and we use the assessment scores to choose between the appearance-based tracker's prediction and the trajectory-based prediction.
Comprehensive evaluations show that the proposed method sets a new state-of-the-art performance on commonly used tracking benchmarks.
\end{abstract}

}

\maketitle

\IEEEdisplaynontitleabstractindextext

%
\IEEEpeerreviewmaketitle

\IEEEraisesectionheading{\section{Introduction}\label{sec:introduction}}

\IEEEPARstart{V}{isual} object tracking is the task of estimating the location of a target object in each frame of an input video, while the initial location of the target is given in the first frame~\cite{smeulders2013visual,yilmaz2006object,yang2011recent}. Recently, a Siamese learning paradigm~\cite{bertinetto2016fully,wang2019SiamMask,li2018high_SiamRPN,zhu2018distractor_DaSiamRPN} has drawn attention in the field.
These approaches cast the visual object tracking problem as learning a general similarity function by computing a cross-correlation between two image regions. Tracking is then performed by finding the feature representations learned for the search region most similar to those of the target-object template. As a result, the visual tracking problem can be solved by Siamese trackers using an end-to-end learning solution.

Despite their relative success, most current approaches including Siamese trackers tend to fail when occlusion occurs, and are erroneous when  objects with similar appearance to the target object are present~\cite{kristan2018sixth_vot2018,wu2015object}. We observe that most current trackers focus on developing discriminative target object’s feature representations~\cite{zhu2018distractor_DaSiamRPN}, online appearance update mechanism\cite{bhat2019DiMP} or  re-detection \cite{wang2018reliable}. A good feature representation or a good online update mechanism is important, however it can be still problematic when target is severely occluded. And, re-detection is just a partial remedy, i.e., useful only when the target object reappears from occlusion.

\begin{figure}[h]
	\begin{center}
		\includegraphics[width=1.0\linewidth]{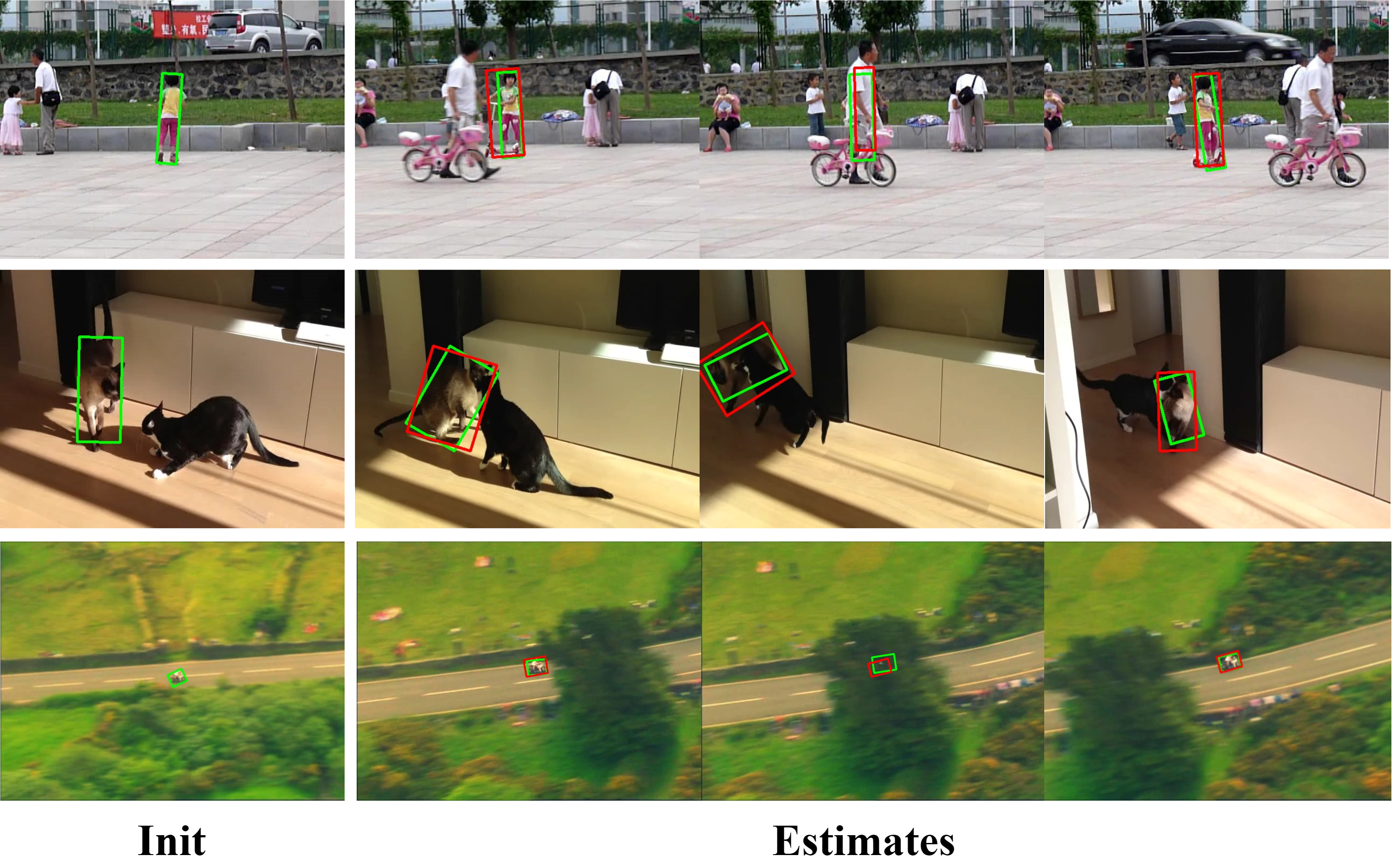}
	\end{center}
	\caption{Our proposed method aims to deal with occlusion by predicting the trajectory. Example frames are shown from the girl (top row), cat (middle row) and road (bottom row) sequences. When the target object suffers from severe/total occlusion, our method (red) is able to predict the target object's locations robustly (green box is ground truth).}
	\label{fig:trailer}
\end{figure}

To address the occlusion problem, we aim at leveraging the predicted future trajectory.
Unlike most existing methods, our basic idea is that, when the target object is severely/totally occluded, the future locations predicted from the motion trajectory  are more reliable than the estimation of the appearance-based tracker. Thus, the future trajectory should be used instead of the tracker's estimation. Based on this idea, we develop a trajectory-guided end-to-end network that predicts the target object's possible locations in the future frames.

To realize the idea, the following are the insights of our algorithm.
First, camera motion significantly affects the target object's locations in the input frames, and should be taken into account.
Second, once camera motion is compensated, the target object's past trajectory represents the target object's motion independent from the camera motion.
Third, previous frames provide the appearance, motion, and surrounding information of the target object, which can contribute to reliable trajectory prediction.

Based on our insights, we design our deep network consisting of 4 subnetworks: an appearance-based tracking network (tracker), a background-motion network, a trajectory network, and  an assessment network,
The tracker estimates the target object's locations solely from appearance information.
It works robustly particularly when the appearance information in the input frame is sufficient and matches to the template.  In our method, any appearance-based tracker from the state-of-the-art method can be used.
The second network in our design, the background-motion network, estimate the camera motion. This camera motion is needed in our method, since the target object's motion in the input video not only depends on the object motion in the world alone but also the camera motion. By estimating and then compensating the camera motion, we can obtain the target object's trajectory independent from the camera motion.
Our third network, the trajectory network, predicts the target object's current and future locations from the target's past observations.

Since our method produces two estimations for a single target object in one frame: one from the tracker and the other from the trajectory network, we need to choose one of them. For this, our fourth network, the assessment network, is trained to assess the accuracy of a prediction. As a result, we can have the confidence score for the tracker prediction and the confidence score for the trajectory prediction. We select the prediction that has higher confidence score. This allows our method to switch dynamically between the tracker and the trajectory prediction, depending on the severity of the occlusion. Note that, to make our confidence scores more reliable, we apply confidence score calibration~\cite{guo2017calibration}.
Figure~\ref{fig:trailer} shows our tracking results even when the target objects are totally occluded.

The summary of our contributions is as follows:
\begin{itemize}
	\item We introduce a background-motion  network that captures the global background motion between consecutive frames to represent the camera motion. This background motion is important to make the target object motion trajectory independent from camera motion.

	\item We propose a trajectory network that learns from the target object’s observations in several previous frames and predicts the locations of the target object in the subsequent future frames. A multi-stream conv-LSTM architecture is introduced to encode and decode the temporal changes of the target object locations in the past.

	\item We present a unified tracking model that integrates background motion modeling, appearance-based tracking, and trajectory prediction to solve the occlusion problem. This unified tracking model can be trained end-to-end.

	\item We introduce a tracking selection mechanism based on the tracker and  trajectory confidence scores, which enables our tracking method to switch automatically and dynamically between the appearance-based tracker and our trajectory-based prediction, particularly when occlusion occurs.
\end{itemize}

The large part of this paper has been published in ECCV 2020 \cite{ECCV2020}. Compared with our ECCV 2020 paper, new contents in this paper includes: unifying the background motion modeling, tracking and trajectory prediction into one deep network (Section 3). This unified model can be trained end-to-end, which further enhances the performance of
our trajectory prediction. Second, to handle overconfident scores of the trajectory confidence score, we employ confidence calibration in the trajectory-guided tracking
mechanism (Section 3.4), which can lead to a significant performance gain. Moreover, we provide the details of our implementation, as well as more comprehensive analysis and evaluation.

\section{Related Works}
\noindent{\bf Visual Object Tracking}
Visual tracking has witnessed a rapid boost in recent years, with the development of a variety of approaches. Recently, the approaches based on Siamese networks\cite{bertinetto2016fully,wang2019SiamMask,li2018high_SiamRPN,zhu2018distractor_DaSiamRPN} have received significant attentions due to their well-balanced tracking accuracy and efficiency. Instead of learning a discriminative classifier online, these approaches formulate visual tracking as a cross-correlation problem and can easily be trained end-to-end using pairs of annotated images. Based on this idea, a number of methods enhance the tracking performance by making use of region proposals \cite{ren2015faster_RPN}, hard negative mining \cite{zhu2018distractor_DaSiamRPN} and binary segmentation \cite{wang2019SiamMask}. A few recent works\cite{yang2017recurrent,cui2016recurrently,yang2018learning,gan2015first,ebrahimi2017ratm} aim to employ temporal information for better object feature representation. Yang \textit{et al.} in \cite{yang2017recurrent} estimate an object-specific filter for tracking by using a Recurrent Neural Networks(RNN). Cui \textit{et al.} in \cite{cui2016recurrently} propose a multi-directional RNN to capture long-range contextual cues. Yang \textit{et al.} in \cite{yang2018learning} propose a dynamic memory network to update the feature representation model. \cite{gan2015first} utilizes a gated recurrent unit to model the temporal information of sequences, while \cite{ebrahimi2017ratm} embeds an attention mechanism into the RNN to help searching target.

Most modern trackers focus on modelling the object feature representation to improve tracking performance. However, relying only on object feature representation can be problematic in the case where the target is occluded or the target comes across other objects with similar appearance. A powerful model update strategy is an effective way to discriminate the similar objects from the target or to handle some partial occlusion cases. However, occlusion is still challenging for these methods. To handle occlusion, some trackers utilize the failure re-detection strategy to search target when the target object appears again after occlusion. However, failure re-detection cannot help to track the
target during occlusion period. To address the occlusion problem completely, we combine the object representation in the spatial domain with the trajectory prediction in the temporal domain by using a spatio-temporal network to track target accurately and robustly.

\begin{figure*}[t]
\centering
\includegraphics[width=1.0\linewidth]{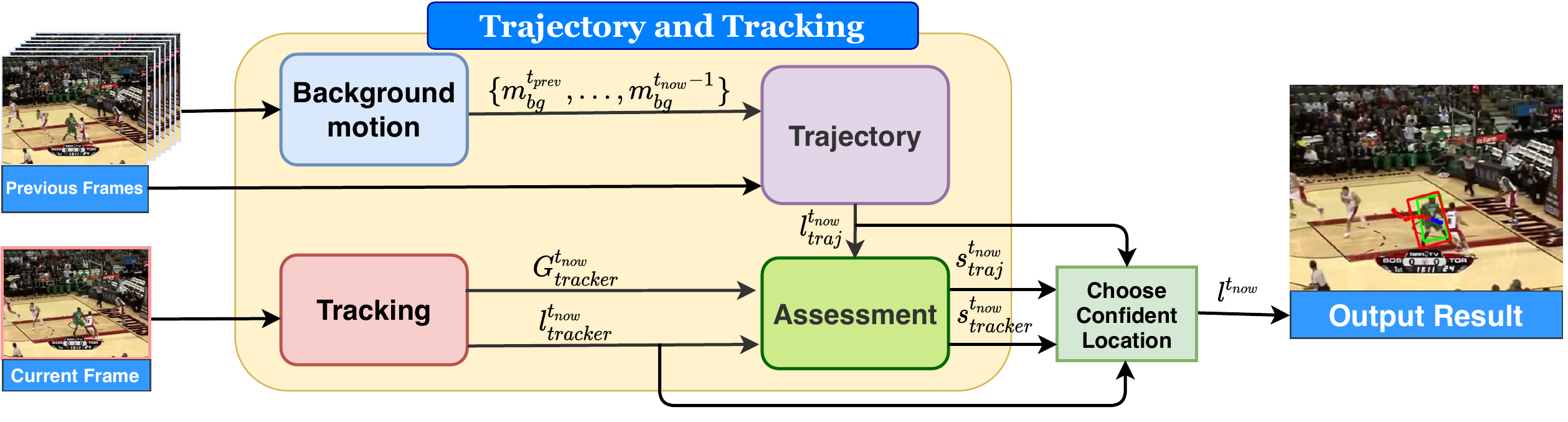}
\caption{Overview of our tracking method. Our approach consists of 4 networks: tracking network (tracker), a background-motion network, a trajectory network, and a assessment network. The tracking network (tracker) estimates the target location $ l^{t_{now}}_{tracker} $ from appearance inference. The background-motion network captures the global background motion vectors $\{ m_{bg}^{t_{prev}}, \dots,  m_{bg}^{t_{now}-1},\}$ to compensate target's trajectory. The trajectory network predicts target's curent and future locations $\{ l^{t_{now}}_{traj}, \dots, l^{t_{future}}_{traj} \}$ from its past observations. The assessment network assess the confidence scores $ s^{t_{now}}_{tracker} $ and $ s^{t_{now}}_{traj} $of $ l^{t_{now}}_{tracker} $ and $ l^{t_{now}}_{traj} $, respectively. The final result $ l^{t_{now}} $  is selected by comparing the calibrated confidence scores.}
\label{fig:overview}
\end{figure*}

\vspace{0.3cm}

\noindent{\bf Trajectory Prediction} Unlike the visual object tracking task, trajectory prediction aims at predicting the target's possible positions in future frames \cite{alahi2016social}. Recently, a large body of works focus on person trajectory prediction by considering human social interactions and behaviors in crowded scenes. Zou \textit{et al.} in \cite{zou2018understanding} learn human behaviors in crowds by using a decision-making process. Liang \textit{et al.} in \cite{liang2019peeking} utilize rich visual features about human behavioral information and interaction with their surroundings to predict future path jointly with future activities. A number of methods try to learn the effects of the physical scene. Scene-LSTM \cite{manh2018scene} divides the static scene into Manhattan grid and predict pedestrian’s location using LSTM. SoPhie \cite{sadeghian2019sophie} combines deep-net features from a scene semantic segmentation model and generative adversarial network using attention to model person trajectories.

There are some methods that take the motion prediction into account for tracking or predicting person path. Amir \textit{et al.} in \cite{sadeghian2017tracking} propose a structure of RNN that jointly reasons on multiple cues over a temporal window for multi-target tracking. Ellis \textit{et al.} in \cite{ellis2009modelling} propose a Gaussian process regression model for pedestrian motion. Hogg \textit{et al.} in \cite{johnson1996learning} propose a statistical model of object trajectories which is learned from image sequences. Compared with these methods, which assume a static camera in modeling the trajectory, our idea is to integrate trajectory prediction into object tracking task using deep learning for a dynamic camera. To simplify the trajectory complexity, several methods split motion into camera motion and object motion. In particular, Takuma \textit{et al.} in \cite{yagi2018future} recently propose an accurate method that makes use of camera ego-motion, scales, the speed of the target person, and the person pose to predict person’s location in future frames. \cite{chandra2019robusttp} proposes an end-to-end trajectory prediction method using an LSTM-CNN network that models the interactions between road-agents in dense traffic. Unlike these methods that make full use of the object surroundings, which are expensive for general single object tracking, we utilize only the past trajectory and target visual features to predict short-term future locations for assisting the tracker to handle occlusion.

\section{Proposed Method}

We develop a tracking method that addresses the occlusion problem.
As shown in Figure~\ref{fig:overview}, our network consists of 4 subnetworks: an appearance-based tracking network (tracker), a background-motion network, a trajectory network, and an assessment network.
Given the current frame at $t_{now}$, the tracker estimates the target object's location $l^{t_{now}}_{tracker}$ based on its appearance.
For severely occluded target object, this $l^{t_{now}}_{tracker}$  is likely to be erroneous.
Our idea is to provide an alternative location using trajectory.
To enable our predicted trajectory to be independent from the camera motion, the background-motion network estimate the global background motion in the previous frames.
Based on the background motion, the target object's previous locations, and the object appearance and surrounding in a few previous frames, our trajectory network predicts the target object's current and future locations $\{l^{t_{now}}_{traj}, l^{t_{now}+1}_{traj}, \dots, l^{t_{future}}_{traj}\} $.
As a result, for the current frame at $t_{now}$, we have two predictions $l^{t_{now}}_{tracker}$ and $l^{t_{now}}_{traj}$.
We create a network that can assess how good the tracking prediction, given the previous tracking predictions, and other information, so that we can assess whether $l^{t_{now}}_{tracker}$ is better than  $l^{t_{now}}_{traj}$, or vice versa. To make the confidence score more accurate, we calibrate the confidence scores  using the temperature scaling technique ~\cite{guo2017calibration}.
Note that, we can employ any existing appearance-based tracking method for our tracker. By employing an existing pre-trained tracker, our unified tracker, background-motion network, trajectory network and assessment network can be trained end-to-end on sequences of annotated data.

\begin{figure*}[t]
\begin{center}
\includegraphics[width=1.0\linewidth]{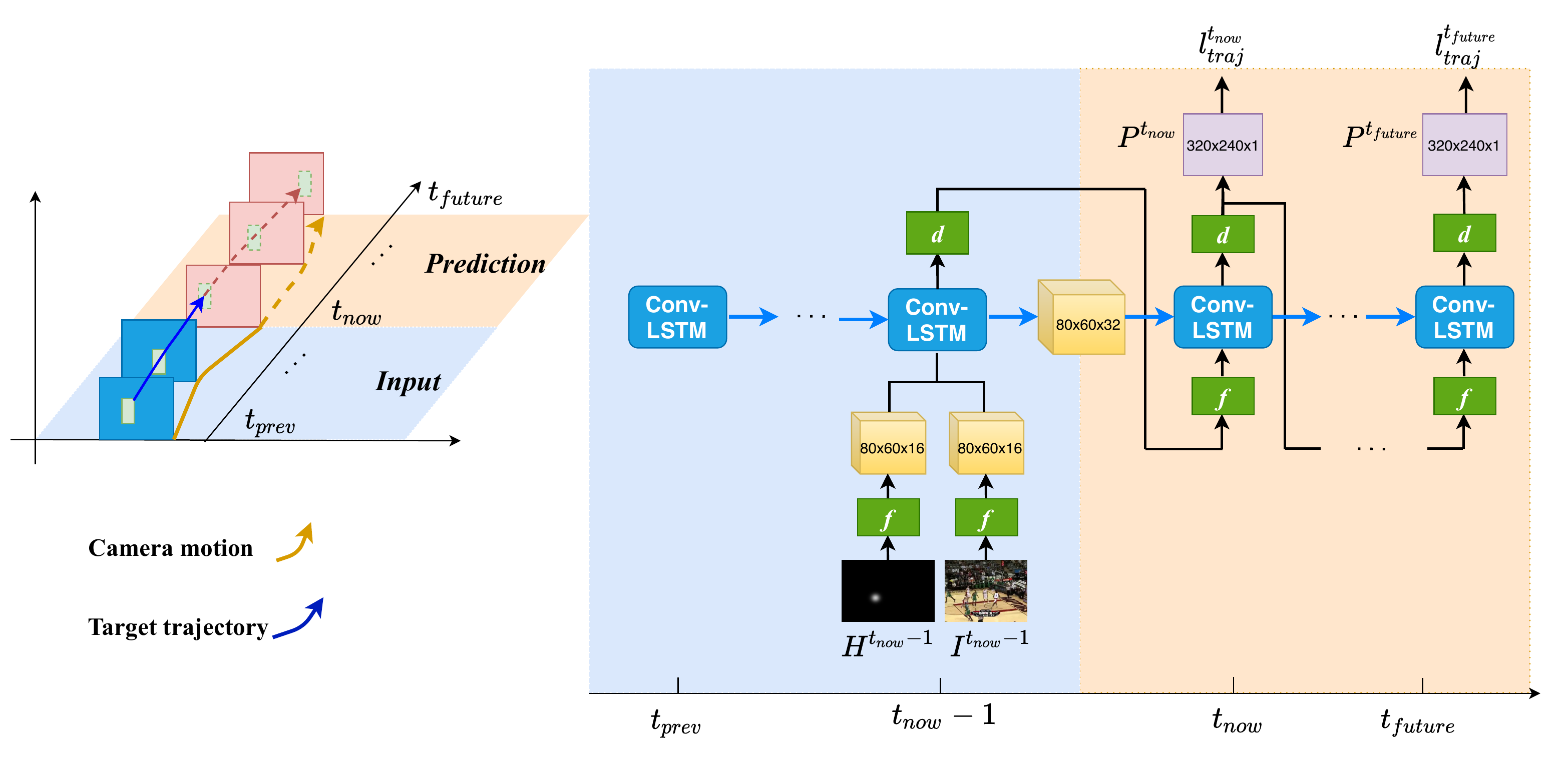}
\end{center}
   \caption{Architecture of our trajectory network. Given  frames at $t_{prev}$ to $t_{now}-1$ as input, we predict the current  at  and future locations of a target object,  at $\{ t_{now},  t_{now}+1, \dots,  t_{future} \}$. Before encoding of Conv-LSTM, a two-stream convolutional network extract features from target's past observations: location maps \{$H^{t_{prev}}, \dots, H^{t_{now}-1} $\} and the corresponding frames \{$ I^{t_{prev}}, \dots, I^{t_{now}-1} $\}. The response maps \{$P^{t_{now}}, \dots, P^{t_{future}}$\} are the output of Conv-LSTM. $ f $ denotes the convolutional module, $ d $ denotes the deconvolutional module.}
\label{fig:pipeline}
\end{figure*}

\subsection{Tracker}
We use a popular Siamese network\cite{bertinetto2016fully} for our appearance-based tracker.
Let $ l^{t_{now}} \in \mathbb{R}^2 $ be the predicted 2D location of the target object at the current frame at $t_{now}$ . Given the initial location $ l^{t_0} $ of a target object  in the first frame of the input video,  the tracker's task is to estimate the target object's location in the current frame at $t_{now}$.
By comparing the target template patch obtained from the initial frame at $t_0$, the tracker produces a similarity map or heatmap $G^{t_{now}}_{tracker}$, from which the predicted location in  the current frame $ l^{t_{now}}_{tracker} $ can be obtained (i.e., from the heatmap's peak).

Heatmap $G_{tracker}$ is produced by applying:
\begin{eqnarray}
G^{t_{now}}_{tracker}=f_{tracker}(x^{t_{now}})*f_{tracker}(z^{t_0}),
\label{eq:tracker}
\end{eqnarray}
where, $ z^{t_0} $  is the target template patch cropped  from the initial frame  at $t_0$. $ x^{t_{now}} $ is a larger patch cropped from the current frame at $t_{now}$ centered on the last estimated location  $l^{t_{now}-1} $. The operator $*$ denotes the cross-correlation operation. $f_{tracker}$ represents the tracking network (the tracker).
While the tracker can work properly when the target object is visible, unfortunately it is erroneous when the target object is severely or totally occluded in a number of consecutive frames. This is because the target object's appearance is severely degraded or disappear, and thus the tracker is unable to find any useful visual information of the target object from the current frame.

To address this occlusion problem, our basic idea is that the occluded target object’s location can be predicted more reliably using the object's past trajectory information. As illustrated in Figure~\ref{fig:pipeline}, we aim at predicting the target object’s locations in the current frame at $t_{now}$ to a frame at $ t_{future} $ (the red boxes the Figure~\ref{fig:pipeline}), namely: $L^{future}_{traj}=\{l^{t}_{traj},\dots,l^{t_{future}}_{traj} \} $, based on observed locations $L^{prev}=\{l^{t_{prev}},\dots,l^{t-1}\} $ in the previous frames (the blue boxes).

\begin{figure}
	\begin{center}
		\includegraphics[width=0.7\linewidth]{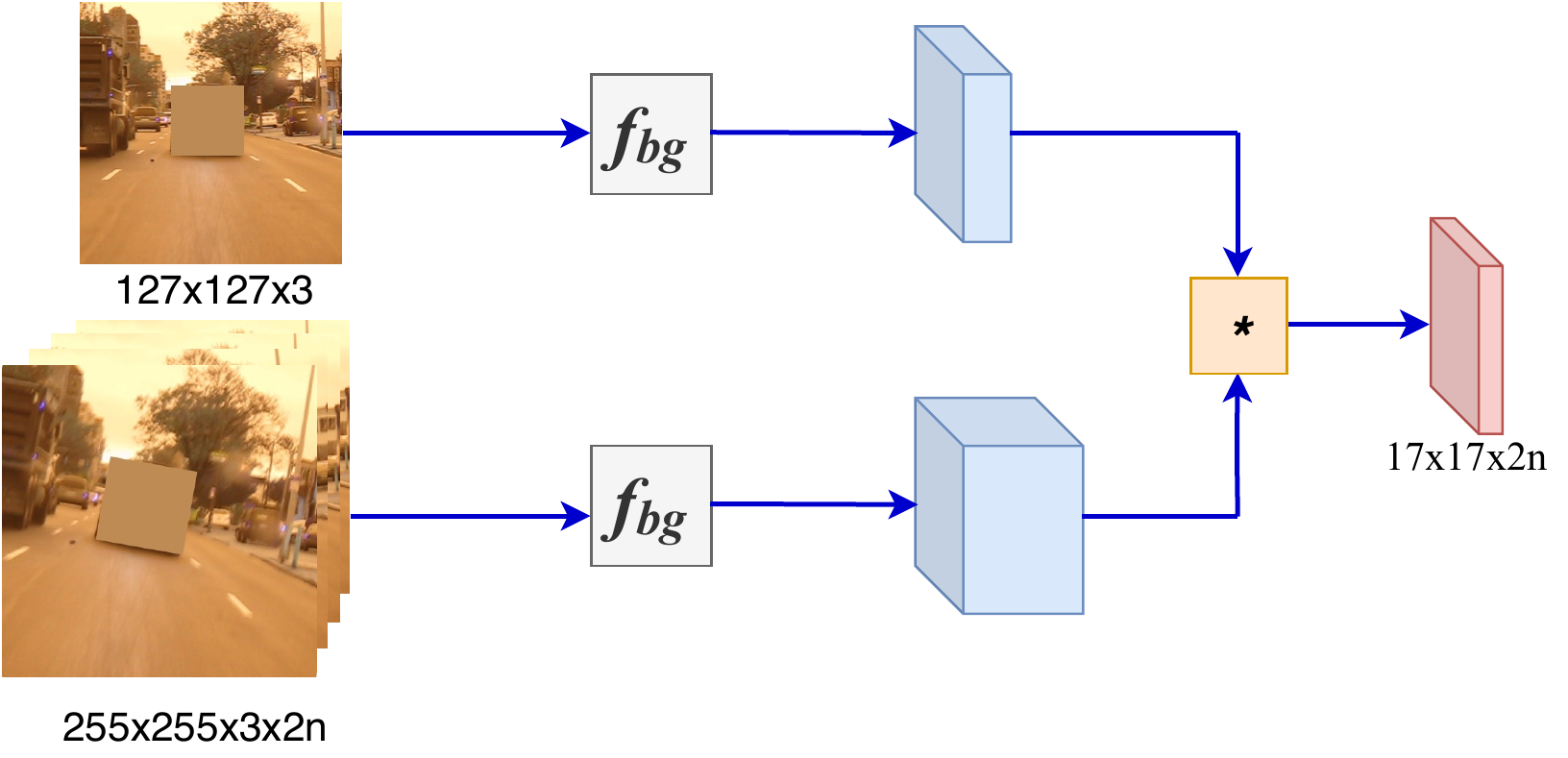}
	\end{center}
	\caption{The architecture of background motion prediction model. We utilize the Siamese network to compare the global background between the adjacent frames. $f_{bg}$ denotes the convolutional network embeding, and $ * $ denotes the cross-correlation. To estimate the scale and rotation changing, we search $2n $ image patches by using patch pyramid with different scale and rotation factors.}
	\label{fig:background_motion}
\end{figure}

\subsection{Background Motion Network}

Predicting $L^{future}_{traj}$  directly from the tracker's estimation $ L^{prev}_{tracker}$ is problematic due to significant camera motion present in the input video. More specifically, the coordinate system representing each location $l^t$ changes dynamically as the camera moves; since the target object's motion appearing on the input sequence  depends on both the object motion in the world and the camera motion.
To address the problem, the camera motion should be estimated and compensated, so that the object motion and thus the object trajectory can be independent from the camera motion.

Generally, the camera motion between adjacent frames, can be represented by rotation and translation. Rotation is described by rotation matrix $R^t \in \mathbb{R}^{3 \times 3}$ and translation is described by vector $V^t\in \mathbb{R}^{3}$, both from a frame at $ t-1 $ to a frame at $t$.
To efficiently estimate the camera motion parameters, we cast the camera motion estimation as matching the global background between two consecutive frames.
Specifically, the 3-D rotation matrix $R^t$ is simplified to one rotation angle $ {r_t}\in \mathbb{R}^{2\times2} $ in the 2-D image domain, and the translation vector $V_t$ is split to the translation vector $ {v_t}\in \mathbb{R}^{2}$, and scale changing, $ c^t$.

To detect the global background motion, we employ a Siamese network that compares the adjacent frames as shown in Figure~\ref{fig:background_motion}.
As the Siamese network is a robust method for image matching, it also can be used to match global background motion.
Unlike the Siamese network for the tracker, the Siamese network here compares the similarity of the global background scenes instead of the target object.
Thus, the patch $ z^{t-1} $ is cropped at the center of the region in the frame $t-1$, and the search image $ x^t$ is the larger cropped image patch in the frame $ t $. To avoid the interference of the target object, the region is masked with one value, i.e., the average value of the whole image region.
The heatmap $G_{bg}$ of background matching in the frame $t$ can be produced by applying:
\begin{equation}
G_{bg}^{t_{now}}= f_{bg}(x^{t_{now}})*f_{bg}(z^{t_{now}-1}),
\label{eq:bg_net}
\end{equation}
where $f_{bg}$ represents the background-motion network.

To be able to handle multiple scales and rotations, we search image patches $ {x_{c_{1}},...,x_{c_{n}}} $ and $ {x_{r_{1}},...,x_{r_{n}},} $ by using patch pyramid with different scale factors $ {c_{1},...,c_{n}} $ and rotation factors $ {r_{1},...,r_{n}} $.
The background motion translation vector $ v^{t_{now}} $ is achieved by finding the displacement of the maximum value and the center of the heatmap. The scale change $ c^{t_{now}} $ and the rotation change $ r^{t_{now}} $ are estimated by searching the position of the maximum value in the multi-scale heatmap set and the multi-rotation heatmap set, respectively.

\begin{figure*}
\begin{center}
\includegraphics[width=1.0\linewidth]{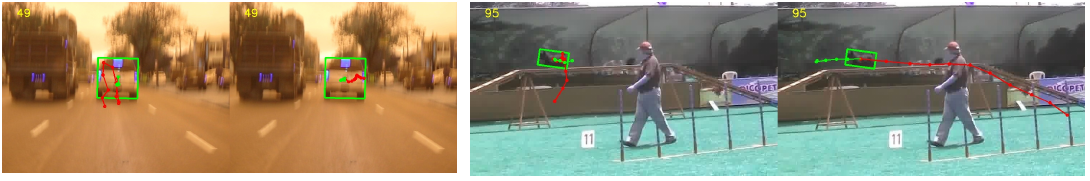}
\end{center}
   \caption{Examples of the target object's original trajectory and  the trajectory that compensates background motion. In each example, the left image is the target's original trajectory and the right one is the one with background motion correction. The red line is the target object's past trajectory and the green line is the target object's future trajectory.}
\label{fig:result}
\end{figure*}

To represent the global motion along multiple frames, we accumulate these vectors,  which represent the local movement between two adjacent frames.
Therefore, for each frame within the input interval [$t_{prev}, t_{now}-1$], we apply:
\begin{equation}
m_{bg}^t=\left\{\begin{array}{ll} r^t c^t v^t & \left(t=t_{prev} + 1\right)
\\ r^t c^t v^t + m_{bg}^{t-1} & \left(t>t_{prev} + 1\right)\end{array}\right.,
\label{eq:cam_motion}
\end{equation}
where $m^t_{bg}$ denote the background motion patterns of frame $t$ in the global background coordinate system. $t \in [t_{prev} , t_{now}-1]$. $ v^t $, $ r^t $ and $ c^t $ denotes the translation, rotation and scale change between two adjacent frames at $t-1$ and $t$.

As shown in Figure~\ref{fig:result}, having compensated the background/camera motion, the trajectories follow the target objects independently from the camera motion. It also shows the Siamese method is able to match global background robustly, even when some moving foreground objects are present.

\subsection{Trajectory Network}
The appearance-based tracker is erroneous when the appearance of the target object is degraded/occluded severely. To address this problem, we rely on the target object's past observations to predict the current most possible location, which also implies predicting a future location.
The straightforward way to predict future locations of the target object is to utilize its previous immediate location (i.e., the temporal smoothness constraint). However, it is insufficient for the object tracking task, since the target object's motion can be arbitrary and the occlusion occurs in a few consecutive frames.

To predict future locations, we develop a fully-convolutional deep network that utilizes a multi-stream conv-LSTM architecture shown in Figure~\ref{fig:pipeline}, which is expressed as:
\begin{equation}
P^t =d(y_{lstm}^{t-1}),
\label{eq:p}
\end{equation}
where:
\begin{eqnarray}
\nonumber
t&\in&[t_{now},t_{future}]\\ \nonumber
y_{lstm}^k &=&LSTM(x_{lstm}^k,y_{lstm}^{k-1}), \\ \nonumber
x_{lstm}^k &=&concat(f_{\mu}(H^k),f_{\theta}(I^k)), \\ \nonumber
 k&\in&[t_{prev},t_{now}-1] \nonumber
\end{eqnarray}
$P^t$ is the response map of a  frame at time $t$. Operator $concat$ represents the concatenation operation of the feature maps. $LSTM$ indicates conv-LSTM.  $y_{lstm}^k$ denotes the output of the LSTM cell at frame $k$. $f_{\mu}$ and $f_{\theta}$ represent two convolutional networks with network parameters $\mu$ and $\theta$, respectively. $d$ denotes the deconvolutional network. $I^k$  is the input image at frame $k$, and $H^k$ is its location map. We obtain $H^k$ by generating a Gaussian function on a map where the peak is at $l^k$; the size of the map is the same as the size of the input frame.

Let us denote the annotated ground-truth location of the target object $GT^t$ on a frame at time $t$. The loss function  for the trajectory prediction is an L1-loss over all predicted future frames:
\begin{equation}
\mathcal{L}_{traj}= \sum_{t=t_{now}}^{t_{future}} || P^t - \mathcal{N}(GT^t) ||,
\end{equation}
where $\mathcal{N}$ indicates a Gaussian function, which the mean is at $GT^t$. Consequently, we can obtain the trajectory prediction, $l^{t_{now}}_{traj}$ from the peak of $P^{t_{now}}$. The same applies to the the future predictions, $l^{t_{now}+1}_{traj}, \dots, l^{t_{future}}_{traj}$.

\subsection{Assessment Network}

Up to this stage, for one frame and a single target object, there are two predicted locations in our method: One is from the tracker model, $l^{t_{now}}_{tracker}$, and the other is from the trajectory module, $l^{t_{now}}_{traj}$. Hence, we need to select them so that we can have one predicted location, $l^{t_{now}}$.
To handle this problem,  we create a network that, given the location prediction, can assess how correct a tracking prediction is.

The network can be expressed as:
\begin{equation}
s^{t_{now}}_* = s(l^{t_{now}}_*, L^{prev}+m_{bg}^{prev}, G^{t_{now}}_{tracker}),
\label{eq:conf_score}
\end{equation}
where $s^{t_{now}}_*$ is the confidence score of how good the tracking prediction, $l^{t_{now}}_*$ is. Function $s$ represents the assessment network.  $L^{prev} = \{l^{t_{prev}}, \dots, l^{t-1}\}$, and $m_{bg}^{prev} =  \{m_{bg}^{t_{prev}}, \dots, m_{bg}^{t_{now}-1}\}$. This equation also implies that, given the information $\{L^{prev}, m_{bg}^{prev}, G^{t_{now}}_{tracker}\}$,  the network predicts how correct the trajectory prediction $l^{t_{now}}_*$ is.

Note that, $l^{t_{now}}_*$ can represent either $l^{t_{now}}_{tracker}$ or $l^{t_{now}}_{traj}$; implying the network can be used to assess the prediction from the tracker and the trajectory prediction, generating two scores: $s^{t_{now}}_{tracker}$ and $s^{t_{now}}_{traj}$. Once we obtained $s^{t_{now}}_{tracker}$ and $s^{t_{now}}_{traj}$, then we can compare them, and then select which tracking prediction is more reliable. This will enable our method to switch automatically and dynamically between the tracker and trajectory predictions.

To train the assessment network, $s$, we rely on  a binary cross entropy:
\begin{equation}
\mathcal{L}_{s}= a \log (s^{t_{now}}_*) +(1-a)\log (1-s^{t_{now}}_*)),
\end{equation}
where $ a\in\{1,0\}$ is the ground-truth. $a=1$ means the prediction is correct, and $a=0$ means the prediction is wrong.
The  network is trained on positive and negative samples.
By computing the intersection over union (IoU) between the ground truth box and the bounding box where the center locates $l^{t_{now}}_*$, this sample will be considered as a positive sample; specifically, when the IoU is larger than 0.5. The corresponding negative sample is generated by adding a random drift displacement on the $l^{t_{now}}_*$ of the positive samples.

\vspace{0.5cm}

\noindent{\bf Trajectory Confidence Calibration} Since we rely on the confidence scores to decide the tracking final prediction, we must ensure that the confidence scores are consistent and linear to the prediction accuracy. For this reason, utilizing a calibration technique \cite{guo2017calibration}, we calibrate our confidence scores.

Let  $s^t_{raw}$ is the non-calibrated score of the assessment network $s$. The calibrated score is obtained by applying:
\begin{equation}
s^t_*=\max_i \sigma \left(s^t_{raw} / T\right)^{(i)}.
\label{eq:calibrate}
\end{equation}
In our case, $ k=2 $.
$T$ is the temperature, $T>0$, which is optimized with respect to negative log likelihood (NLL)\cite{friedman2001elements,lecun2015deep} on the validation set. The network’s parameters are fixed during this calibration process.

\section{Implementation Details}

\subsection{Overall Algorithm}

\begin{algorithm}[htb]
	\caption{Tracker, Background Motion and Trajectory}
	\label{alg:overall}
	\begin{algorithmic}[1]
		\REQUIRE ~~\\
		Frame $I^{t_{now}}$, Previous frames $I^{t_{prev}},...,I^{t_{now}-1}$\\
		Previous target locations $L^{prev}=\{l^{t_{prev}},...,l^{t_{now}-1}\}$
		\ENSURE ~~\\
		Target object location $l^{t_{now}}$ in the  current frame\\
		Future object locations $\{l^{t_{now}+1}, \dots, l^{t_{future}}\}$
		\\
		{\bf Tracker Estimation:}\\
		\STATE Compute heatmap $G^{t_{now}}_{tracker}$ using Eq.~(\ref{eq:tracker})
		\STATE Set $l^{t_{now}}_{tracker}$ from  the highest value in $G^{t_{now}}_{tracker}$
		\\
		{\bf Background-Motion Estimation:}\\
		\STATE Compute heatmaps $\{ G^{t_{prev}}_{bg}, \dots, G^{t_{now}-1}_{bg} \}$ using Eq.~(\ref{eq:bg_net})
		\STATE Compute background motion $\{m_{bg}^{t_{prev}}, \dots, m_{bg}^{t_{now}-1} \}$ using Eq.~(\ref{eq:cam_motion})
		\\
		{\bf Trajectory Prediction:}\\
		\STATE Generate location maps $\{H^{t_{prev}},..., H^{t_{now}-1} \}$ based on $\{ m_{bg}^{t_{prev}},..., m_{bg}^{t_{now}} \}$ and $L^{prev}$
		\STATE Compute response maps $\{ P^{t_{now}},...,P^{t_{future}} \}$ using Eq.~(\ref{eq:p})
		\STATE Obtain the trajectory prediction  $l^{t_{now}}_{traj}$ from  $P^{t_{now}}$
		\\
		{\bf Selection:}\\
		\STATE Obtain confidence scores $s^{t_{now}}_{traj}$ and $s^{t_{now}}_{tracker}$ using Eq.(~\ref{eq:conf_score})
		\STATE Calibrate $s^{t_{now}}_{traj}$ and $s^{t_{now}}_{tracker}$ using Eq.~(\ref{eq:calibrate})
		\STATE Set $l^{t_{now}}$  from either $l^{t_{now}}_{traj}$ and $l^{t_{now}}_{tracker}$ depending on $s^{t_{now}}_{traj}$ and $s^{t_{now}}_{tracker}$
		
	\end{algorithmic}
\end{algorithm}

Algorithm~\ref{alg:overall} summarizes our overall method. Given the input frames and previous target locations, we want to estimate the current location, as well as the future locations. To achieve this, at step 1, we compute the heatmap, $G_{tracker}$ from using Eq.~(\ref{eq:tracker}). In this equation, we require the object template, $z^{t-1}$, which can be obtained from the previous tracking prediction, to search the target object in the current frame. In step 2, we obtain the target object's location, $l^{t_{now}}_{tracker}$ from the maximum  value in map $G^{t_{now}}_{tracker}$.

In step 3, we compute the background-motion heatmaps $\{ G^{t_{prev}}_{bg}, \dots, G^{t_{now}-1}_{bg} \}$ using Eq.~(\ref{eq:cam_motion}). To apply this matching equation, we need template
$z^{t_-1}$ and search region $x^t$ of the respective frames. In step 4, we can obtain $\{m_{bg}^{t_{prev}}, \dots, m_{bg}^{t_{now}-1} \}$, which are the background motions from frame $t_{prev}$ to frame $t_{now}-1$.

Once we have the background motion information, we can predict the trajectory. In step 5, we generate the location maps, $\{H^{t_{prev}},..., H^{t_{now}-1} \}$, which each of them is represented by a map containing a Gaussian function whose peak is at the predicted location. These maps are independent from the background/camera motion. In step 6, we compute the response maps, $\{ P^{t_{now}},...,P^{t_{future}} \}$, using our conv-LSTM networks described in Eq.~(\ref{eq:p}). Based on the current frame response map, $P^{t_{now}}$, we can obtain the predicted location of the target object, $l^{t_{now}}_{traj}$.

Hence, we have two predictions: $l^{t_{now}}_{tracker}$ from step 2 and $l^{t_{now}}_{traj}$ from step 7. In step 8, based on Eq.(~\ref{eq:conf_score}), we compute the confidence scores, and in step 9, we calibrate them, producing: $s^{t_{now}}_{tracker}$  and $s^{t_{now}}_{traj}$. Based on the comparison between these confidence scores, in step 10, we choose either $l^{t_{now}}_{tracker}$ or $l^{t_{now}}_{traj}$ as our final tracking prediction for the current frame.

\subsection{Network Architecture}
\begin{figure}
\begin{center}
\includegraphics[width=1.0\linewidth]{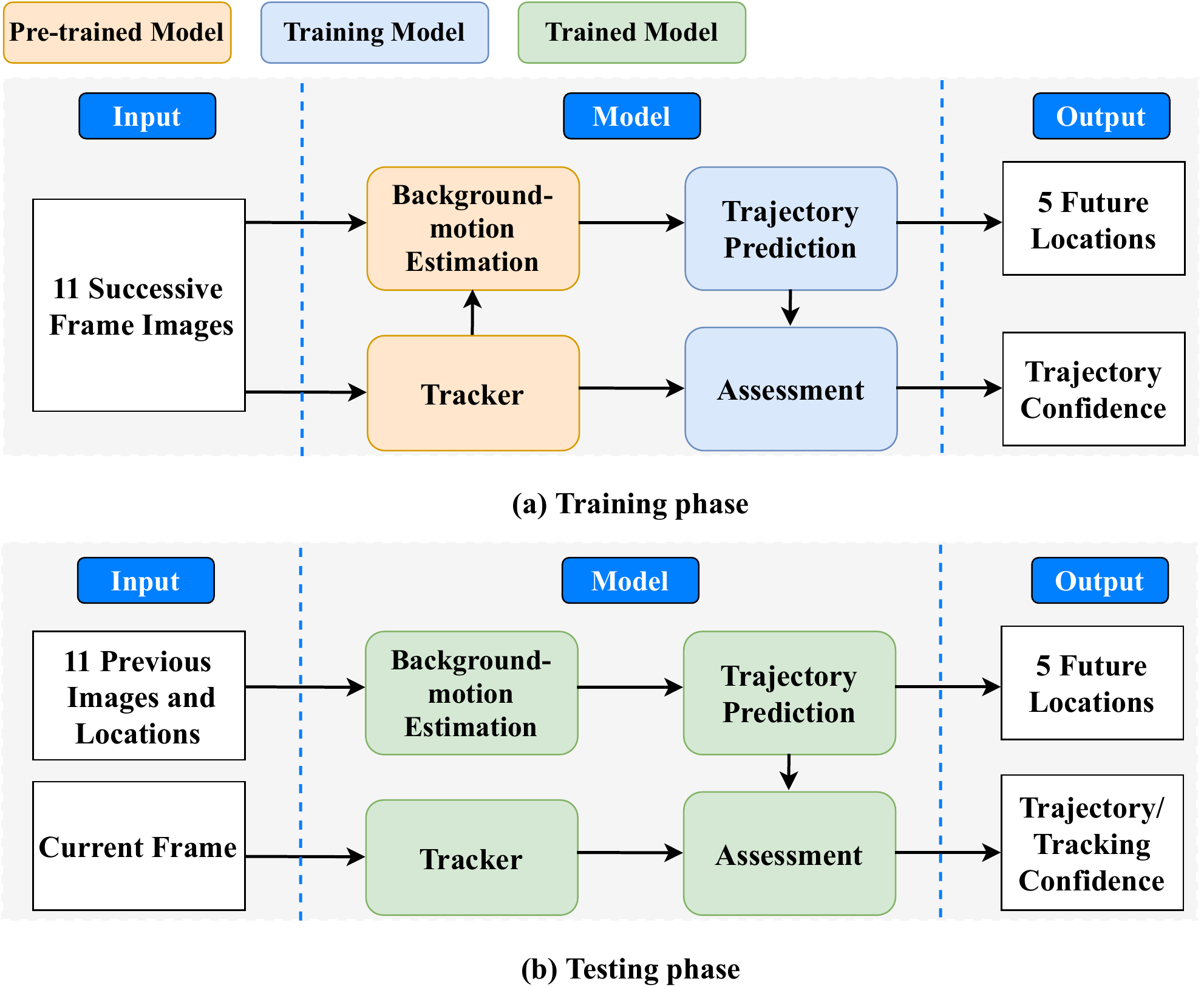}
\end{center}
   \caption{The training and testing pipelines of our proposed method.}
\label{fig:implementation}
\end{figure}

Figure~\ref{fig:implementation} shows the pipelines of the training and testing of our implementation.
We use the target object's observations in the previous 11 frames to predict the locations of the target in next 5 future frames.
For the trajectory network, we use a general seq-to-seq LSTM network\cite{sutskever2014sequence} as our backbone, where the encoder consists of 11 LSTM cells and the decoder consists of 5 LSTM cells.
As the input sequence is a spatio-temporal data, we extend the matrix multiplications of LSTMs to a convolution operation on the multi-channel feature maps, which preserves the spatial information of the target object.

The size of all convolutional filters is $ 3 \times 3 $. A two stream of fully-convolutional network with 5 convolutional layers of stride 4 extracts features from the location map and image. The two sets of feature maps are concatenated  before the LSTM network. After decoding of the LSTM, each response map representing the future location state is generated by three deconvolutional layers with stride 4. Finally, based on the target object's past 11 locations and the tracker's current heatmap score, the confidence score is obtained from a network of 3 1-D convolutional layers followed by a sigmoid activate function. For the background motion prediction network, we choose the popular Siamese network Siamfc\cite{bertinetto2016fully}. For our tracking network, we choose an existing tracking method's network (ie, DiMP\cite{bhat2019DiMP} or SiamMask\cite{wang2019SiamMask}).

\subsection{Training}
For the trajectory network, we choose 16 consecutive frames in the input video as the temporal range of one sample set. For each sample set, we select the first 11 frames as the past states and regard the rest 5 frames as the future states.
For the trajectory classifier network, we choose the target object's locations in first 12 frames from the successive 16 frames as one sample set. For each sample set, the last frame location is the future state. Specifically, we consider data augmentation by random translations (up to the mean of the displacements in the past frames).

During training, the background-motion network and the tracking network use the pre-trained models in Figure 6, which means the parameters of these two networks are fixed. Thus, the input of our unified model is 12 frames sequence (11 previous frames and current frame), while the output is 5 response maps in future 5 frames and a confidence score that represents the trajectory confidence of the trajectory prediction status. Note that, the assessment network only outputs one score, and performs twice as mentioned in Section 3.4. The label of the response map is produced by a Gaussian function. The peak location is based on the target object's location by compensating the background motion vector.
We use the Adam optimization with the fixed learning rate of 0.0002 for 20 epochs. We train our unified model by using ImageNet-VID \cite{russakovsky2015imagenet}.

\subsection{Testing}
During testing, the background-motion network and trajectory network take in 11 previous frames and the target's past locations to predict 5 future locations. The tracking network only process the current frame to estimate the target location and also provide the heatmap. The assessment network outputs the confidence score of the trajectory prediction and tracking status. Note that, our method needs the first 11 frames states. To deal with the initialization problem, we repeat the first frame state to reach the 11 frames initialization quantity at the first 11 frames. It can be seen as the target object stays still at the initial location. Our trajectory obtains the motion information from the tracker's results. For faster processing speed, it is not necessary to predict trajectory every frame until the tracker gets a low confidence score.

\begin{table*}[]
\caption{State-of-the-art comparison on the VOT2019 dataset in terms of expected
average overlap (EAO), accuracy and robustness. Compared with the baseline tracker DIMP, our approach Ours-DIMP achieves improvement on all three metrics and obtains a significant performance gain on the robustness.}
\centering
\begin{tabular}{lcccccc}
\hline
 & \begin{tabular}[c]{@{}c@{}} SiamRPN++ \\ \cite{li2019siamrpn++} \end{tabular}
 & \begin{tabular}[c]{@{}c@{}} DCFST \\ \cite{zheng2019DCFST} \end{tabular}
 & \begin{tabular}[c]{@{}c@{}} ATOM \\ \cite{danelljan2019atom} \end{tabular}
 & \begin{tabular}[c]{@{}c@{}} SiamMask \\ \cite{wang2019SiamMask} \end{tabular}
 & \begin{tabular}[c]{@{}c@{}} DIMP \\ \cite{bhat2019DiMP} \end{tabular}
 & \begin{tabular}[c]{@{}c@{}} {\bf Ours-DiMP} \end{tabular} \\
\hline
EAO $\uparrow$             & 0.285 & 0.361 & 0.292 & 0.287 & 0.305 & 0.323 \\
Accuracy $\uparrow$        & 0.599 & 0.589 & 0.603 & 0.602 & 0.589 & 0.596 \\
Robustness $\downarrow$    & 0.482 & 0.321 & 0.411 & 0.426 & 0.361 & 0.301 \\
\hline
\end{tabular}
\end{table*}

\begin{table*}[]
\scriptsize
\caption{ State-of-the-art comparison on OTB2013, OTB2015 and UAV123 datasets in terms of area-under-the-curve (AUC) score. The best results are shown in bold. Ours-DIMP improves the base tracker DIMP on all three datasets. Especially, our approach achieves the best AUC score on the OTB2015 and UAV123 datasets.}
\centering
\begin{tabular}{lcccccccccc}
\hline
 & \begin{tabular}[c]{@{}c@{}} DeepSTRCF\\\cite{li2018learning}\end{tabular}
 & \begin{tabular}[c]{@{}c@{}} DaSiamRPN\\\cite{zhu2018distractor_DaSiamRPN}\end{tabular}
 & \begin{tabular}[c]{@{}c@{}}ATOM \\ \cite{danelljan2019atom} \end{tabular}
 & \begin{tabular}[c]{@{}c@{}}CCOT \\ \cite{danelljan2016CCOT} \end{tabular}
 & \begin{tabular}[c]{@{}c@{}}MDNet \\ \cite{nam2016MDNet} \end{tabular}
 & \begin{tabular}[c]{@{}c@{}} ECO\\ \cite{danelljan2017eco} \end{tabular}
 & \begin{tabular}[c]{@{}c@{}} SiamRPN++ \\ \cite{li2019siamrpn++} \end{tabular}
 & \begin{tabular}[c]{@{}c@{}}UPDT \\ \cite{bhat2018UPDT} \end{tabular}
 & \begin{tabular}[c]{@{}c@{}}DiMP \\ \cite{bhat2019DiMP} \end{tabular}
 & \begin{tabular}[c]{@{}c@{}}{\bf Ours-DiMP}\end{tabular}
 \\
\hline
OTB2013 & 68.2 & 66.0 & 66.0 & 68.0 & 71.1 & 71.2 & 69.1 & {\bf 71.4} & 69.1 & 69.9 \\
OTB2015 & 67.5 & 66.0 & 66.7 & 67.3 & 67.8 & 69.1 & 69.6 & 69.1 & 67.7 & {\bf 70.1} \\
UAV123  &  -   & 56.9 & 64.3 & 50.7 &  -   & 50.6 & 64.2 & 54.2 & 64.3 & {\bf 65.1}
\\
\hline
\end{tabular}
\end{table*}

\section{Experiment Results}

We evaluate our approach on VOT-2019\cite{kristan2019seventhVOT2019}, OTB2013\cite{smeulders2013visual}, OTB2015\cite{wu2015object} and UAV123\cite{mueller2016UAV123} benchmarks. We choose the Siamese framework based tracker DIMP\cite{bhat2019DiMP} and SiamMask\cite{wang2019SiamMask} as our baseline trackers. On a single Nvidia GTX 1080Ti GPU, we achieve a tracking speed of 11 FPS when employing DIMP as the base tracker and 13 FPS for SiamMask.

In this section, we provide a comprehensive comparison of Ours-DIMP with 12 state-of-the-art methods in the literature. The trackers used for our comparison are: DaSiamRPN\cite{zhu2018distractor_DaSiamRPN}, SiamRPN\cite{li2018high_SiamRPN}, ATOM\cite{danelljan2019atom}, CCOT\cite{danelljan2016CCOT}, MDNet\cite{nam2016MDNet}, ECO\cite{danelljan2017eco}, SiamRPN++\cite{li2019siamrpn++}, UPDT\cite{bhat2018UPDT}, DCFST\cite{zheng2019DCFST}, DeepSTRCF\cite{li2018learning}, SiamMask\cite{wang2019SiamMask} and DIMP\cite{bhat2019DiMP}. Some of these trackers use the Siamese framework as backbone.

\subsection{Evaluation Results}

\begin{figure*}
\begin{center}
\includegraphics[width=1.0\linewidth]{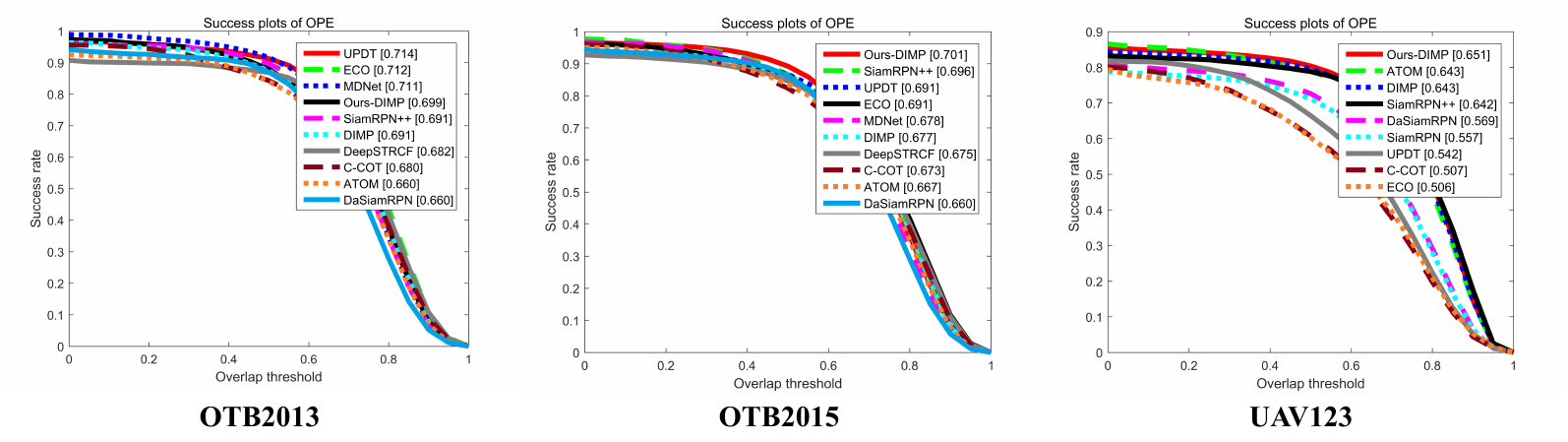}
\end{center}
   \caption{Overlap success plots on the OTB2013, OTB2015 and UAV123 datasets. The legend of overlap success contains area-under-the-curve score for each tracker. The proposed algorithm, Ours-DIMP, performs favorably against the state-of-the-art trackers.}
\label{fig:calibration}
\end{figure*}

\noindent{\bf VOT2019 Dataset\cite{kristan2019seventhVOT2019}:}
We evaluate our approach on the 2019 version of Visual Object Tracking (VOT) consisting of 60 challenging videos. Following the evaluation protocol of VOT2019, we adopt the expected average overlap (EAO), accuracy (average overlap over successfully tracked frames) and robustness (failure rate) to compare different trackers. The detailed comparisons are reported in Table 1. Compared to DIMP, our approach has a $ 6\%$ lower failure rate, while achieving a better accuracy. This shows that trajectory prediction is crucial for robust tracking.

\vspace{0.2cm}

\noindent{\bf OTB2013 Dataset\cite{smeulders2013visual}:}
This dataset consists of 50 videos with different attributes including occlusion. Table 2 shows the AUC scores on the OTB2013 dataset. Among the compared methods, UPDT\cite{bhat2018UPDT} achieves the best results with an AUC score of $71.4\%$. Ours-DiMP achieves an AUC score of $69.9\%$, compared with our baseline tracker DiMP $69.1\%$. We also report overlap success plots on the OTB2013 \cite{smeulders2013visual} in Figure 7.

\vspace{0.2cm}

\noindent{\bf OTB2015 Dataset\cite{wu2015object}:}
Table 2 shows the AUC scores over all the 100 videos in the dataset. Ours-DiMP achieves the best results with an AUC score of $70.1\%$, compared with our baseline tracker DiMP $67.7\%$. We also report overlap success plots on the OTB2015 dataset in Figure 7. Our approach improves the baseline tracker DIMP on both OTB2013 and OTB2015. Compared with our little improvement on OTB2013, our approach obtains a significant performance gain on OTB2015, which contains more occlusion videos.

\vspace{0.2cm}

\noindent{\bf UAV123 Dataset\cite{mueller2016UAV123}:}
This dataset consists of 123 low altitude aerial videos captured from a UAV. Compared to other datasets, UAV123 has heavier camera motion that affects the target’s trajectory severely. Testing on UAV dataset is challenging for our approach based on trajectory prediction. AUC results are shown in Table 2. Ours-DIMP, achieves the best AUC score of $65.1\%$, verifying the strong trajectory prediction abilities of our tracker under the heavy camera motion.

\begin{table*}
\scriptsize
\caption{ Attribute-based comparison with state-of-the-art trackers on the OTB2015 dataset. We report the AUC scores ($\%$) for the state-of-the-art trackers. The number
of videos associated with the attribute is shown in parenthesis. Compared with DIMP, our approach provides improved performance on all 11 attributes, especially occlusion.}
\centering
\begin{adjustbox}{width=1\textwidth}
\begin{tabular}{lccccccccccc}
\hline
 & \begin{tabular}[c]{@{}c@{}} Low \\ resolution (9) \end{tabular}
 & \begin{tabular}[c]{@{}c@{}} Background \\ clutter (31) \end{tabular}
 & \begin{tabular}[c]{@{}c@{}} Out of \\ View (14) \end{tabular}
 & \begin{tabular}[c]{@{}c@{}} Occlusion \\ (49) \end{tabular}
 & \begin{tabular}[c]{@{}c@{}} Motion \\ blur (31) \end{tabular}
 & \begin{tabular}[c]{@{}c@{}} Scale \\ variation (65) \end{tabular}
 & \begin{tabular}[c]{@{}c@{}} Deformation \\ (44) \end{tabular}
 & \begin{tabular}[c]{@{}c@{}} Illumination \\ variation (38) \end{tabular}
 & \begin{tabular}[c]{@{}c@{}} Fast \\ motion (42) \end{tabular}
 & \begin{tabular}[c]{@{}c@{}} Out-of-plane \\ rotation (63) \end{tabular}
 & \begin{tabular}[c]{@{}c@{}} In-plane \\ rotation (51) \end{tabular}
 \\
\hline
{\bf Ours-DIMP}                            & {\bf 63.3} & {\bf 67.3} & {\bf 64.6} & {\bf 68.2} & {\bf 71.9} & {\bf 70.6} & {\bf 68.8} & {\bf 70.9} & {\bf 69.1} & {\bf 67.9} & {\bf 69.0}\\
DIMP\cite{bhat2019DiMP}                    & 58.4 & 62.7 & 60.2 & 63.5 & 69.0 & 67.8 & 65.8 & 68.5 & 67.7 & 66.7 & 68.5\\
\hline
\end{tabular}
\end{adjustbox}
\end{table*}

\begin{table*}
\scriptsize
\caption{ Attribute-based comparison with the baseline tracker on the UAV123 dataset. We report the AUC scores ($\%$). The number of videos associated with the attribute is shown in parenthesis. Compared with DIMP, our approach provides improved performance on 8 of 11 attributes, especially occlusion.}
\centering
\begin{adjustbox}{width=1\textwidth}
\begin{tabular}{lccccccccccc}
\hline
 & \begin{tabular}[c]{@{}c@{}} Low \\ resolution (70) \end{tabular}
 & \begin{tabular}[c]{@{}c@{}} Background \\ clutter (60) \end{tabular}
 & \begin{tabular}[c]{@{}c@{}} Out of \\ View (31) \end{tabular}
 & \begin{tabular}[c]{@{}c@{}} Occlusion \\ (28) \end{tabular}
 & \begin{tabular}[c]{@{}c@{}} Motion \\ blur (73) \end{tabular}
 & \begin{tabular}[c]{@{}c@{}} Scale \\ variation (48) \end{tabular}
 & \begin{tabular}[c]{@{}c@{}} Deformation \\(33) \end{tabular}
 & \begin{tabular}[c]{@{}c@{}} Illumination \\ variation (109) \end{tabular}
 & \begin{tabular}[c]{@{}c@{}} Fast \\ motion (30) \end{tabular}
 & \begin{tabular}[c]{@{}c@{}} Out-of-plane \\ rotation (68) \end{tabular}
 & \begin{tabular}[c]{@{}c@{}} In-plane \\ rotation (21) \end{tabular}
 \\
\hline
{\bf Ours-DIMP}                & {\bf 66.7} & {\bf 64.9} & {\bf 63.5} & {\bf 62.4} & {\bf 58.9} & {\bf 50.8} & 43.5 & {\bf 63.4} & {\bf 59.7} & 61.5 & 47.1\\
DIMP\cite{bhat2019DiMP}        & 65.5 & 64.0 & 62.2 & 60.8 & 57.5 & 49.5 & 43.5 & 62.6 & 58.4 & {\bf 61.6} & {\bf 47.8}\\
\hline
\end{tabular}
\end{adjustbox}
\end{table*}

\subsection{Attribute-based Analysis}

\begin{figure}
\begin{center}
\includegraphics[width=1.0\linewidth]{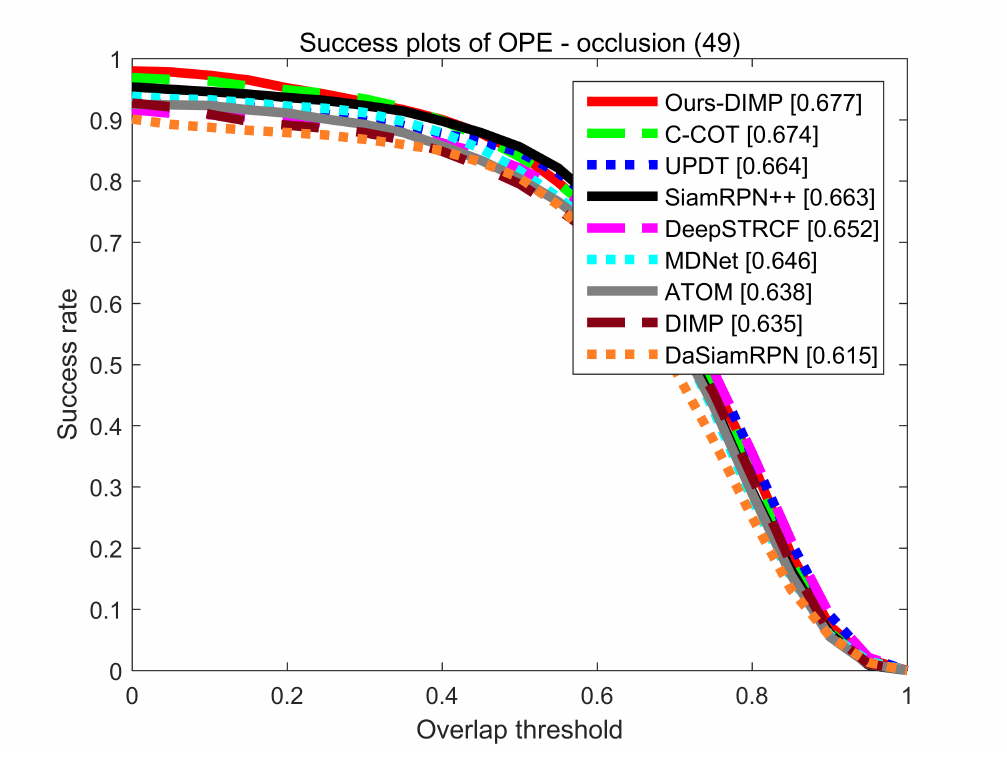}
\end{center}
   \caption{Occlusion attribute comparison with state-of-the-art trackers in terms of AUC score on the OTB2015 dataset. The best results are obtained with Ours-DIMP method, improving the baseline tracker DIMP by $4.2\%$ in AUC.}
\label{fig:calibration1}
\end{figure}

\begin{figure}
\begin{center}
\includegraphics[width=1.0\linewidth]{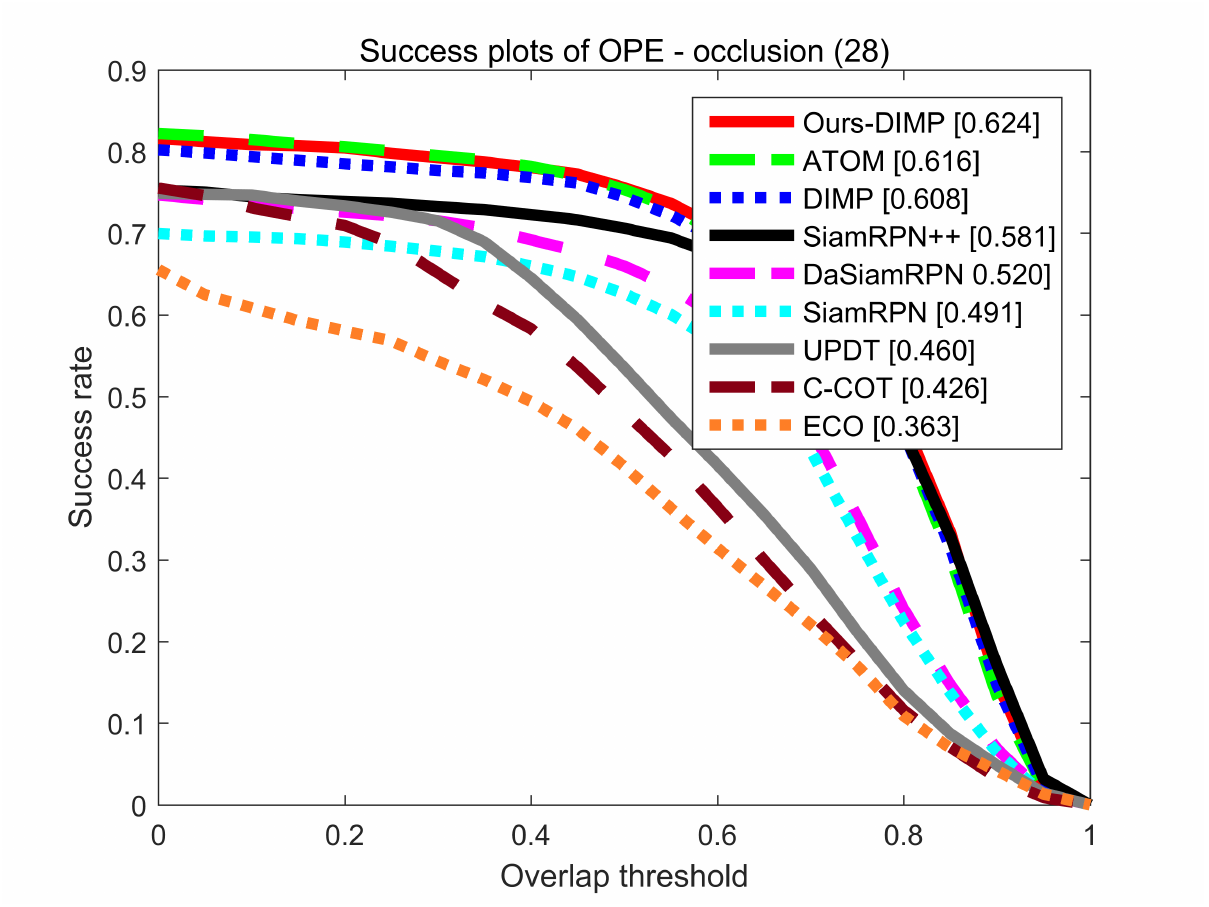}
\end{center}
   \caption{Occlusion attribute comparison with state-of-the-art trackers in terms of AUC score on the UAV123 dataset. The best results are obtained with Ours-DIMP method, improving the baseline tracker DIMP by $1.6\%$ in AUC.}
\label{fig:calibration2}
\end{figure}

To analyze the performance on occlusion and other video attributes, we compare our method with the baseline tracker DiMP on OTB2015 and UAV123 datasets. In the OTB and UAV datasets, all videos are annotated with 11 different attributes, namely: low resolution, background clutters, out-of-view, occlusion, motion blur, scale variation, deformation, illumination variation, fast motion, out-of-plane rotation and inplane rotation.
Table 3 shows the AUC scores of all the 11 attributes in the OTB2015 dataset. Our approach provides favorable results on all 11 attributes compared with DiMP. Specially, our method achieves a significant gain of $4.2\%$ on the occlusion attribute and obtains a improvement of about $4\%$ in AUC score on low resolution, background clutter, out of view attributes. Our method achieves the best AUC score on occlusion attribute in Figure 8. The AUC scores of the 11 attributes in the UAV123 dataset are reported in Table 4. Our method outperforms DIMP with a relative gain of $1.6\%$ on the occlusion attribute. Our method also achieves the best AUC score on occlusion attribute in Figure 9.

Since the occlusion and out of view attributes diminish the target object’s appearance in the image frame, it is challenging for the appearance-based tracker to detect target in the current image. For the low resolution and motion blur attributes degrade the target object’s appearance, our method also performs better than baseline tracker. For rotation and fast motion situation that have little interference on object’s appearance, our method obtains similar performance compared with DiMP's. These results demonstrate the effectiveness of our trajectory
prediction.

\subsection{Ablation Studies}
We perform an analysis on the proposed model prediction architecture. Experiments are performed on VOT2018\cite{kristan2018sixth_vot2018} dataset.

\vspace{0.2cm}

\noindent{\bf Impact of Different Baseline Trackers}
Since the baseline tracker provides our method the heatmap score and tracking state, its performance is important to our method. To analyze the influence of baseline tracker, we choose another popular Siamese-framework based tracker SiamMask \cite{wang2019SiamMask} as our baseline tracker. Compared with SiamMask, our method, namely Ours-SiamMask, improves the performance on all EAO, accuracy and robustness criteria. In particular, Ours-SiamMask obtains a significant relative gain of $5.7\%$ in EAO, compared to SiamMask. Compared with the baseline variant tracker SiamMask-LD which improved by training on larger dataset, our corresponding, namely Ours-SiamMask-LD, achieves the gain of $2\%$ and $8.4\%$ in EAO and robustness respectively. Our method obtains a processing speed of 13 FPS which includes the processing time of the baseline tracker. The results for this analysis verifies the strong generalization abilities of our method, as shown in Table 5.

\vspace{0.2cm}

\begin{table*}[]
\caption{Analysis of different tracker models on the VOT2018 dataset in terms of EAO, accuracy and robustness. Compared with SiamMask/SiamMask-LD, our approach obtains a significant improvement on all three metrics, especially the robustness.}
\centering
\begin{tabular}{lcccc}
\hline
 & \begin{tabular}[c]{@{}c@{}} SiamMask\cite{wang2019SiamMask} \end{tabular}
 & \begin{tabular}[c]{@{}c@{}} SiamMask-LD\cite{wang2019SiamMask} \end{tabular}
 & \begin{tabular}[c]{@{}c@{}} {\bf Ours-SiamMask} \end{tabular}
 & \begin{tabular}[c]{@{}c@{}} {\bf Ours-SiamMask-LD} \end{tabular} \\
\hline
EAO $\uparrow$             & 0.380 & 0.422 & 0.402 & {\bf 0.442 } \\
Accuracy $\uparrow$        & 0.610 & 0.599 & {\bf 0.618 } & 0.607 \\
Robustness $\downarrow$    & 0.281 & 0.234 & 0.249 & {\bf 0.197 } \\
Speed $\uparrow$           & {\bf 60 } & 43 & 13 & 13 \\
\hline
\end{tabular}
\end{table*}

\noindent{\bf Impact of Multi-Cues}
We make an ablation study to see how the background motion cue, the location map cue and image cue contribute overall tracking performances respectively. We compare three different inputs. \textbf{No bg:} The trajectory prediction network predicts the target object’s locations without background motion compensation. Thus, the camera motion will affects the target object’s trajectory. \textbf{No img:} Here, we use only the location map cue and the background motion cue. \textbf{No loc:} We utilize the target object’s location value directly instead of the location map. The results are shown in Table 6. \textbf{No bg} achieves an EAO score of 0.394, even worse than the baseline tracker. \textbf{No img}, which can exploit background information, provides a substantial improvement, achieving an EAO score of 0.439. This highlights the importance of employing the background motion prediction compensation. Our complete method  outperforms \textbf{No loc} by $0.7\%$. This shows that the location map is a better way to represent the target object’s trajectory in the image domain. Our complete method obtains the best results, which means each cue in our inputs is beneficial to improve performance.

\vspace{0.2cm}

\begin{table}[!t]
\renewcommand{\arraystretch}{1.3}
\caption{Analysis of the impact of multi-cues on the VOT2018 dataset. The complete approach performs best, which means each cue is beneficial to improve the ability of trajectory prediction.}
\centering
\begin{tabular}{lccccc}
\hline
 & \begin{tabular}[c]{@{}c@{}} Tracker \\ \end{tabular}
 & \begin{tabular}[c]{@{}c@{}} No bg \\ \end{tabular}
 & \begin{tabular}[c]{@{}c@{}} No img \\ \end{tabular}
 & \begin{tabular}[c]{@{}c@{}} No loc \\ \end{tabular}
 & \begin{tabular}[c]{@{}c@{}} Ours \end{tabular} \\
\hline
EAO $\uparrow$   & 0.422 & 0.394 & 0.439 & 0.435 & {\bf 0.442 } \\
\hline
\end{tabular}
\end{table}

\begin{table}[!t]
\renewcommand{\arraystretch}{1.3}
\caption{Analysis of the impact of different temporal range of inputs on the VOT2018 dataset. Compared with longer time window of inputs, the shorter one performs better as the target's motion is varying with time.}
\centering
\begin{tabular}{lcccc}
\hline
 & \begin{tabular}[c]{@{}c@{}} Tracker \\ \end{tabular}
 & \begin{tabular}[c]{@{}c@{}} Temp-51 \\ \end{tabular}
 & \begin{tabular}[c]{@{}c@{}} Temp-21 \\ \end{tabular}
 & \begin{tabular}[c]{@{}c@{}} Ours \end{tabular} \\
\hline
EAO $\uparrow$   & 0.422 & 0.426 & 0.437 & {\bf 0.442 } \\
\hline
\end{tabular}
\end{table}

\noindent{\bf Impact of Temporal Range}
We analyze the impact of the input’s temporal range. Our basic idea is that using a short-term time window of one or two seconds for observation to predict the target object’s future location. Thus, we choose 21 frames and 51 frames as our temporal ranges, namely Temp-21 and Temp-51 respectively. Our complete method, namely Ours, takes 11 frames, as mentioned in section 4. The results are reported in Table 7. The Temp-21 variant outperforms the Temp-51 variant by $1.1\%$. Our complete method with the temporal range of 11 frames obtains the best performance. This indicates that the method with shorter length of the temporal range obtains better performance. This is due to the target motion is vary with time and also relative to the testing sequences.

\vspace{0.2cm}

\noindent{\bf Impact of Selection Mechanism}
We analyze the impact of trajectory selection mechanism by comparing three different variants. \textbf{No heatmap:} the selection score is evaluated based on the target trajectory without the heatmap score. \textbf{Weight:} The heatmap score is treated as a weight to the confidence score instead of the trajectory classifier network’s input. \textbf{Ours:} A sub-classifier network takes in target’s locations and corresponding heatmap score, and outputs a confidence score, as described in section 3.4. The results are shown in Table 8. \textbf{No heatmap} even makes the baseline tracker worse. In contrast, by considering the heatmap score from the tracker’s appearance inference, our method obtains a significant gain of about $2\%$ in EAO score over the baseline tracker. It also indicates that combing the heatmap score into a network is a more effective way than using it as a weight. These results demonstrate that our method can effectively switches between the tracker state and trajectory prediction.

\vspace{0.2cm}

\begin{table}[!t]
\renewcommand{\arraystretch}{1.3}
\caption{Analysis of the impact of different selection mechanisms on the VOT2018 dataset. By using the heatmap score from tracker, the selection mechanism significantly improves the performance of the tracker.}
\centering
\begin{tabular}{lcccc}
\hline
 & \begin{tabular}[c]{@{}c@{}} Tracker \\ \end{tabular}
 & \begin{tabular}[c]{@{}c@{}} No Heatmap \\ \end{tabular}
 & \begin{tabular}[c]{@{}c@{}} Weight \\ \end{tabular}
 & \begin{tabular}[c]{@{}c@{}} Ours \end{tabular} \\
\hline
EAO $\uparrow$   & 0.422 & 0.326 & 0.434 & {\bf 0.442 } \\
\hline
\end{tabular}
\end{table}

\begin{figure}
\begin{center}
\includegraphics[width=1.0\linewidth]{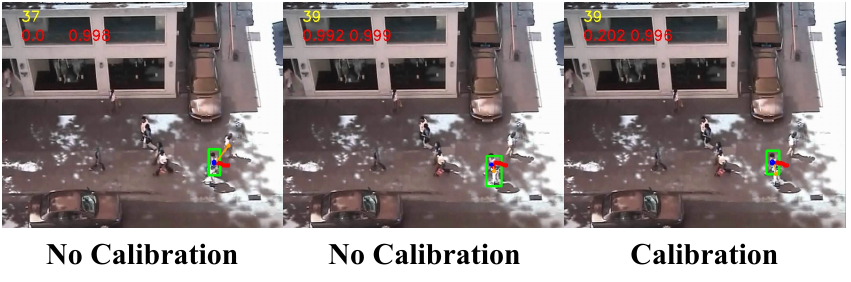}
\end{center}
   \caption{A comparison example of adding trajectory prediction and calibrating trajectory confidence. In each image, the number at top-left denotes the image frame number, the two numbers in the second row denote the trajectory confidence of the tracking result and trajectory predicting result. The red dots denote the target’s past locations, the blue ones are our prediction, and orange one is the tracking result.}
\label{fig:calibration}
\end{figure}

\begin{figure*}
\begin{center}
\includegraphics[width=1.0\linewidth]{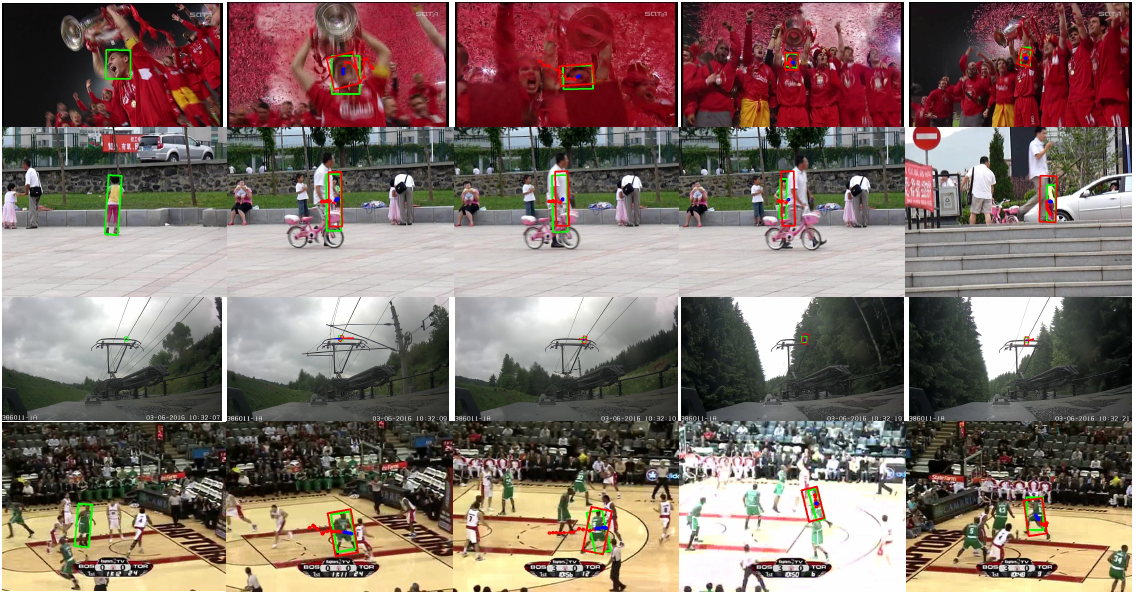}
\end{center}
   \caption{Qualitative results of Ours-SiamMask for sequences from VOT2018. The sequence names, from top to bottom, are soccer1, girl, conduction1, and basketball. The red dots denote the target’s past locations and the blue ones are our prediction. In comparison with the ground truth (green bounding box), our method(red one) performs well under full occlusion in Girl and Soccer1 sequences.}
\label{fig:vot18}
\end{figure*}

\noindent{\bf Impact of Confidence Calibration and Unify Model}
We analyze the impact of Confidence Calibration and unified model by comparing three different variants. \textbf{No Calibration:} the final result is selected by using trajectory confidence score without calibration. \textbf{No Unify:} The background motion modeling, appearance-based tracking, and trajectory prediction are trained and tested independently. \textbf{Ours:} Our complete method unifies the model and calibrates the trajectory confidence. The results are shown in Table 9. Our complete method obtains the best results, which means calibrating confidence and unifying model are beneficial to improve performance. Furthermore, a qualitative comparison example is shown in Figure 10. In the left image in Figure 10, when the baseline tracker tracks drift due to the similar appearance object nearby, our approach assistant tracker to find the right target object by comparing their trajectory confidence. However, when the wrong tracking result is close to the target’s past trajectory in the middle image of Figure 10, the trajectory confidence is confused to classify. By calibrating confidence, it can be solved in the right image of Figure 10.

\begin{table}[!t]
\renewcommand{\arraystretch}{1.3}
\caption{Analysis of the impact of Confidence calibration and Unify model on the VOT2018 dataset. The complete approach, namely Ours, obtains the best performance, which means the calibrating confidence and unifying model are beneficial to improve performance.}
\centering
\begin{tabular}{lcccc}
\hline
 & \begin{tabular}[c]{@{}c@{}} Tracker \\ \end{tabular}
 & \begin{tabular}[c]{@{}c@{}} No calibration \\ \end{tabular}
 & \begin{tabular}[c]{@{}c@{}} No unify \\ \end{tabular}
 & \begin{tabular}[c]{@{}c@{}} Ours \end{tabular} \\
\hline
EAO $\uparrow$   & 0.422 & 0.437 & 0.440 & {\bf 0.442 } \\
\hline
\end{tabular}
\end{table}

\subsection{Qualitative Evaluation}
Here we provide a qualitative comparison of our approach. Figure 11 illustrates frames from four sequences with occlusion (girl and soccer) and significant camera motion (girl, soccer, conduction1 and basketball). In the girl sequence, the target girl is totally hidden in occlusion. It is difficult to find ant useful visual information in the image. By predicting future locations from the target's past trajectory, our approach prevents the tracker drift. In the soccer sequence, the background is clutter and the target is partly occluded. In the conduction1 sequence, the unobvious appearance of the target is challenging for the general appearance-based tracker. In the basketball sequence, it is also difficult for tracking due to the target motion changing dramatically with surrounding and the similar objects nearby. These cases can be solved by our approach using trajectory prediction in Figure 11. All these four sequences contains significant camera motion, our approach performs well compared with ground truth, while the trajectory is also shown smooth in Figure 11. This shows the strong compensating trajectory ability of our background-motion estimation. Again, our approach demonstrates robustness in these scenarios and is able to keep tracking the target throughout the sequence.

\section{Conclusions}
We introduce a novel tracking method, which is trained in end-to-end, and combines background motion modeling, appearance-based tracking, and trajectory prediction to solve the occlusion problem. We present a background motion model that captures the global background motion between adjacent frames to the effect of camera motion. We propose a new trajectory prediction model that learns from the target object’s observations in several previous frames and predicts the locations of the target object in the subsequent future frames. A multi-stream conv-LSTM architecture is introduced to encode and decode temporal evolution in these observations. We also present a trajectory-guided tracking mechanism by using a calibrated trajectory confidence score, which helps the tracker to switch dynamically between the current tracker prediction and our trajectory prediction, particularly when the target object is occluded.

We conduct extensive experiments on the Online Tracking Benchmark (OTB), the Unmanned Aerial Vehicle (UAV) and the Visual Object Tracking (VOT) 2019, 2018 challenge datasets. The results clearly demonstrate that our approach provides significant improvement over the baseline trackers DIMP~\cite{bhat2019DiMP} and SiamMask~\cite{wang2019SiamMask}, especially under occlusion. We also compare our approach with several state-of-the-art trackers. Our method outperforms 8 state-of-the-art trackers in the literature on the OTB and UAV datasets. For the occlusion attribute on the OTB and UAV datasets, our tracking approach is shown to outperform 8 state-of-the-art trackers, obtaining the top ranking AUC score.

\ifCLASSOPTIONcompsoc
  \section*{Acknowledgments}
\else
  \section*{Acknowledgment}
\fi

Robby T. Tan’s research in this work is supported by the National Research Foundation, Singapore under its Strategic Capability Research Centres Funding Initiative.

\ifCLASSOPTIONcaptionsoff
  \newpage
\fi



\bibliographystyle{IEEEtran}
\bibliography{IEEEabrv,reference}





%

\begin{IEEEbiography}[{\includegraphics[width=1in,height=1.25in,clip,keepaspectratio]{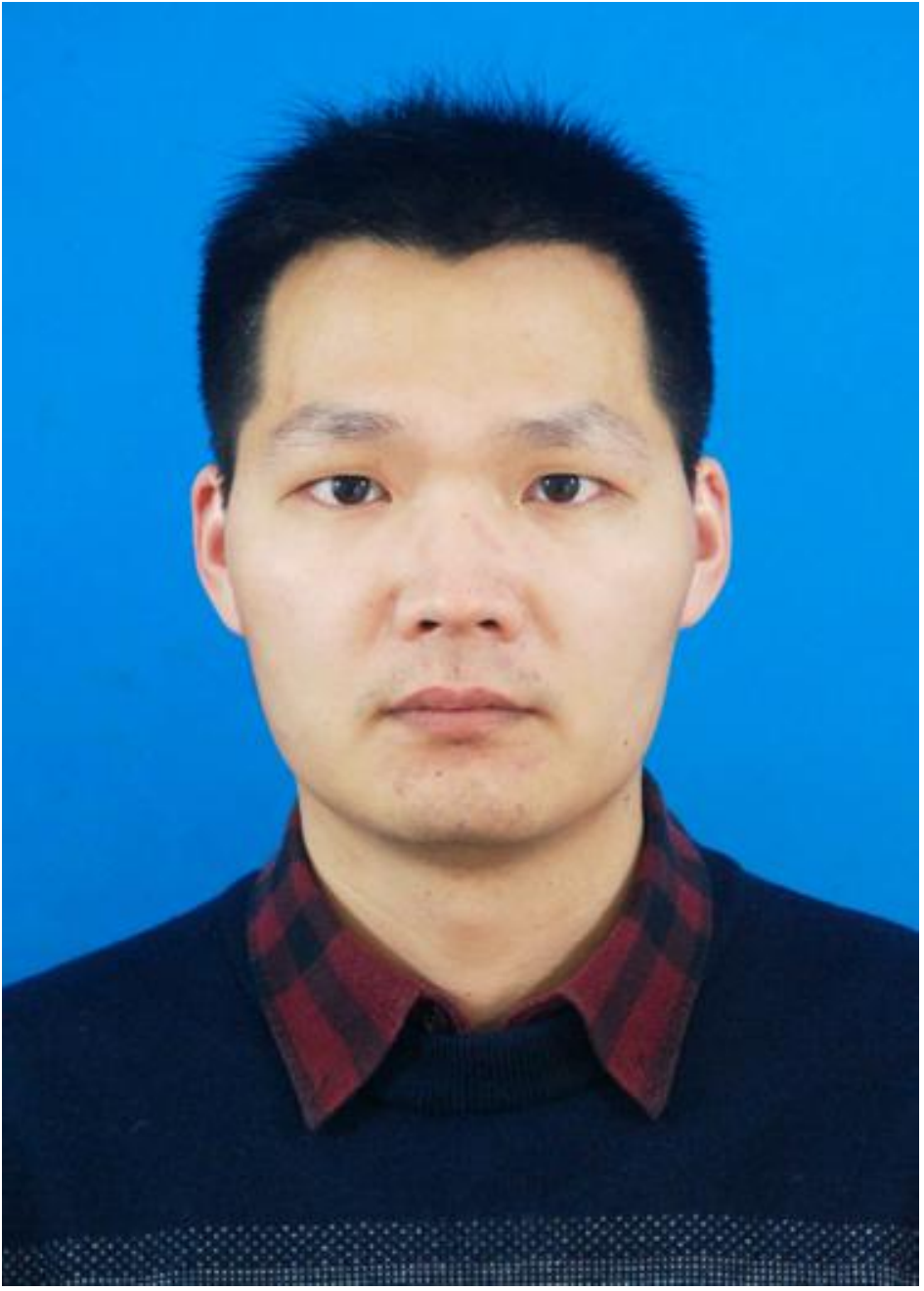}}]{Yuan Liu} received the BS degree in Optical Engineering from Nanjing University of Science and Technology, China, in 2012, and currently working towards Ph.D. degree in Nanjing University of Science and Technology. He was a visiting student with the National University of Singapore, from 2019 to 2020.  His current research interests include visual object tracking and deep learning.
\end{IEEEbiography}

\begin{IEEEbiography}[{\includegraphics[width=1in,height=1.25in,clip,keepaspectratio]{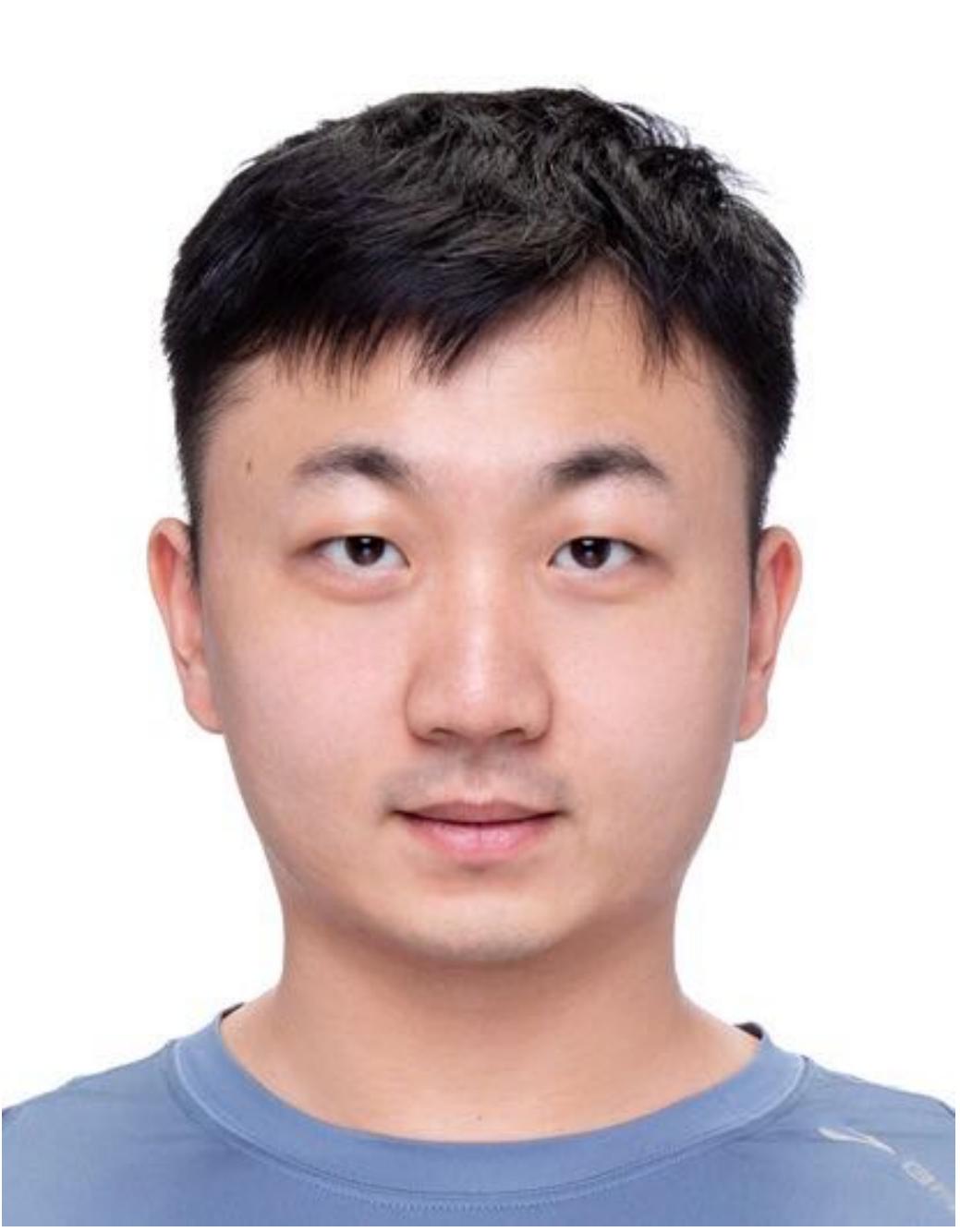}}]{Ruoteng Li}  received the B.S. degree from National University of Singapore, Singapore, in 2013 and received his Ph.D degree from electrical and computer engineering, National University of Singapore in 2020. His research direction includes deep learning based image processing (deraining, dehazing, denoising, etc.), optical flow, stereo estimation, depth estimation and their related fields.
\end{IEEEbiography}

\begin{IEEEbiography}[{\includegraphics[width=1in,height=1.25in,clip,keepaspectratio]{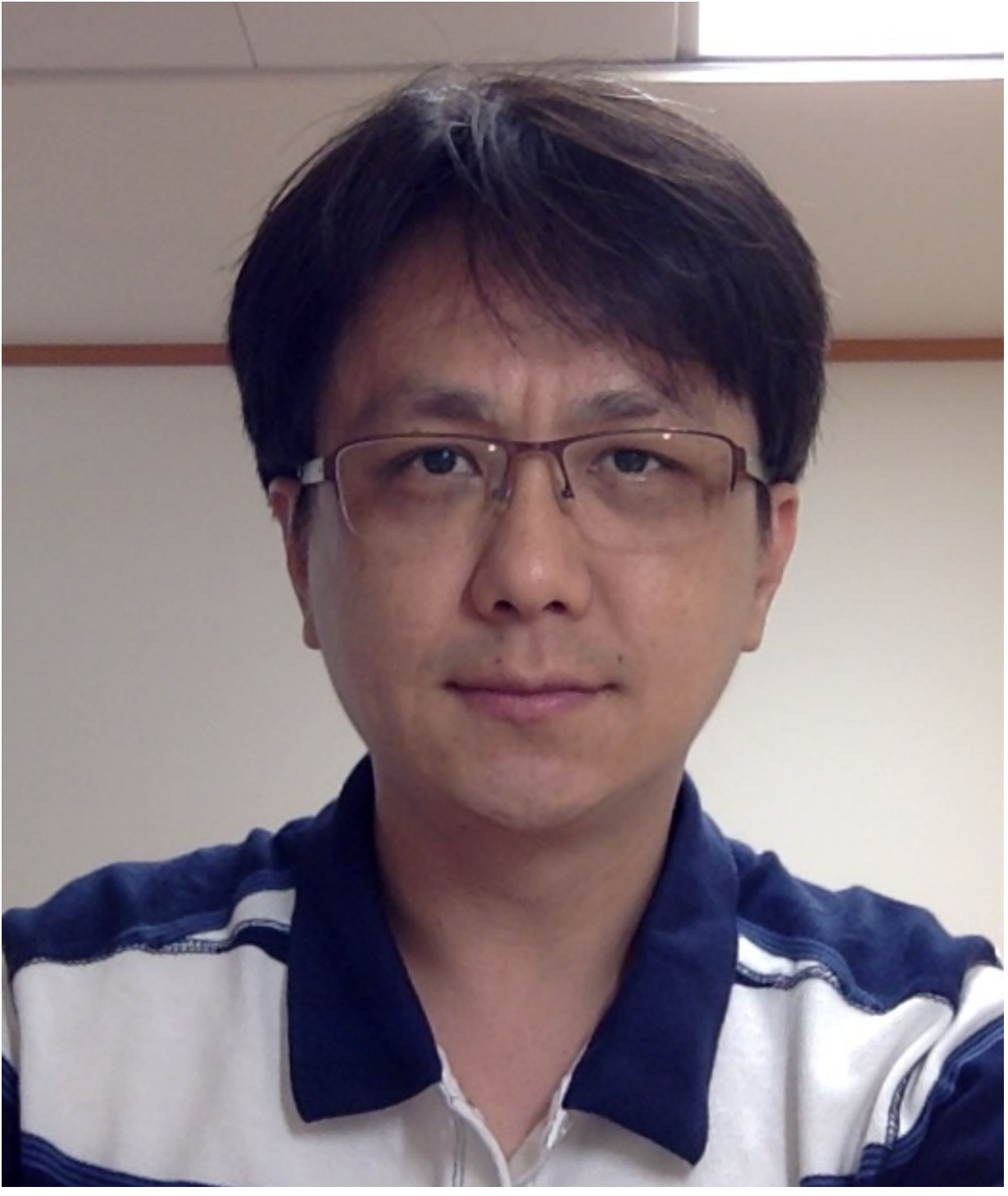}}]{Robby T. Tan} received the PhD degree in computer science from the University of Tokyo. He is
an associate professor at both Yale-NUS College
and ECE (Electrical and Computing Engineering),
National University of Singapore. Previously, he
was an assistant professor at Utrecht University.
His research interests include computer vision
and deep learning. He is a member of the IEEE.
\end{IEEEbiography}

\begin{IEEEbiography}[{\includegraphics[width=1in,height=1.25in,clip,keepaspectratio]{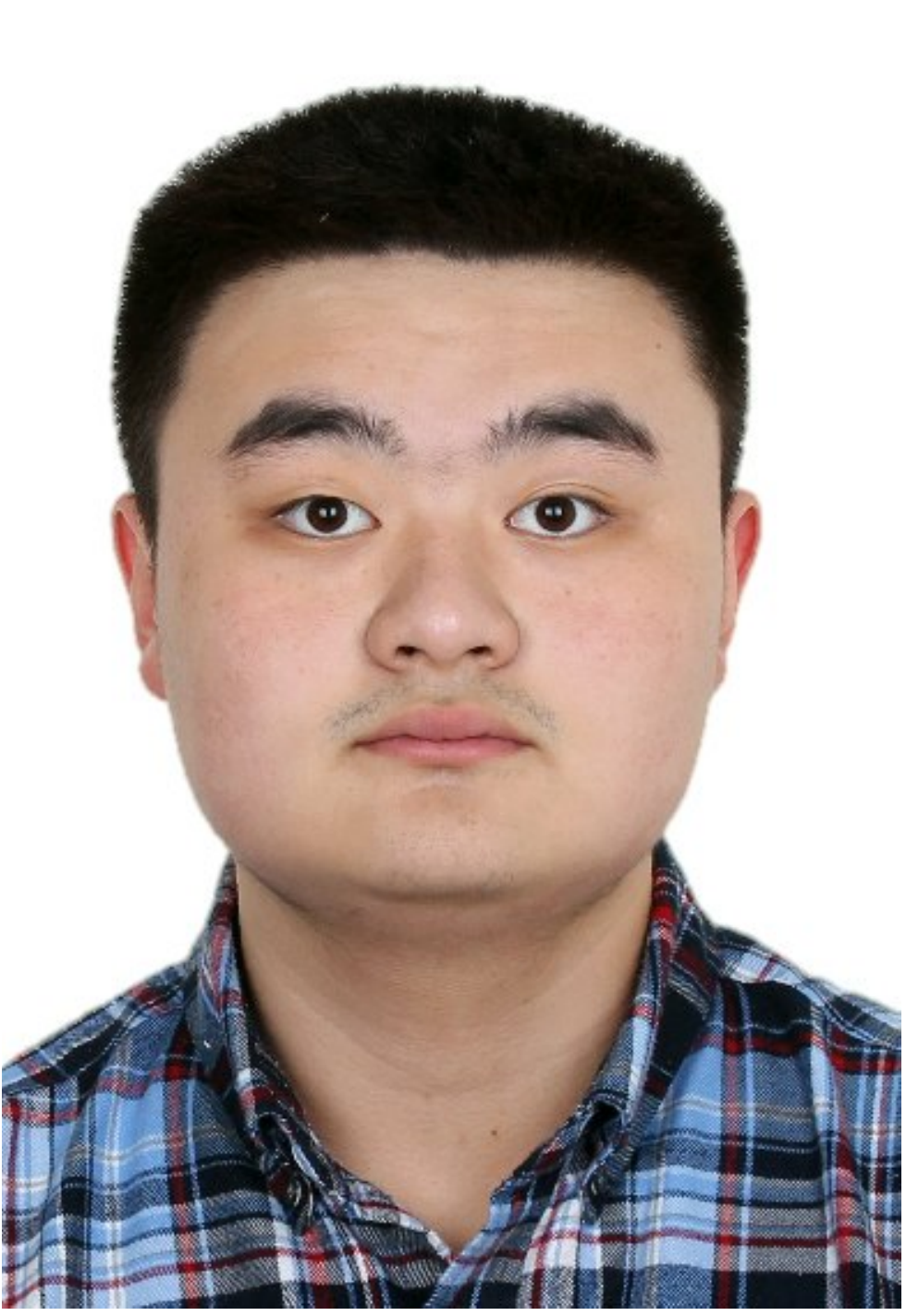}}]{Yu Cheng}
 received the BS degree in Electric and Electronic Engineering from Nanyang Technological University, Singapore, in 2018, and currently working towards Ph.D. degree in National University of Singapore. His current research interests
include deep-learning based human pose tracking, 2D/3D human pose estimation, face recognition and related applications.
\end{IEEEbiography}

\begin{IEEEbiography}[{\includegraphics[width=1in,height=1.25in,clip,keepaspectratio]{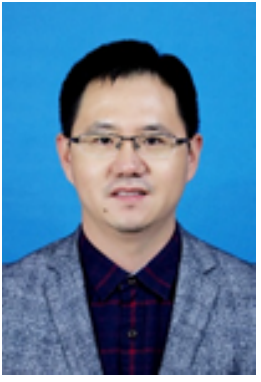}}]{Xiubao Sui}
received the PhD degree in Optical Engineering from Nanjing University of Science and Technology, China.  He is currently a professor with the School of Electronic and Optical Engineering, Nanjing University of Science and Technology, China. His current research interests include deep-learning based image processing, image denoising, super-resolution reconstruction and related applications. He is a member of the IEEE.
\end{IEEEbiography}









\end{document}